\documentclass[lettersize,journal]{IEEEtran}
\usepackage{amsmath,amsfonts,amssymb,amsthm}
\usepackage{algorithmic}
\usepackage{algorithm}
\usepackage{booktabs}
\usepackage{multicol}
\usepackage{multirow}
\usepackage{booktabs}
\usepackage{tcolorbox}
\usepackage{enumerate}
\usepackage{xcolor}
\usepackage{colortbl}
\usepackage{tabularray}
\usepackage{microtype}
\usepackage[utf8]{inputenc}
\usepackage[T1]{fontenc}
\UseTblrLibrary{booktabs}
\usepackage{tabularx}
\usepackage{hhline}
\usepackage{paralist}
\usepackage{lipsum}
\usepackage{array}
\usepackage[caption=false,font=scriptsize,labelfont=sf,textfont=sf]{subfig}
\usepackage{textcomp}
\usepackage{stfloats}
\usepackage{soul}
\usepackage{url}
\usepackage{verbatim}
\usepackage{graphicx}
\usepackage{hhline}
\usepackage{xcolor}
\usepackage{cite}
\usepackage[colorlinks,
linkcolor=black,
anchorcolor=black,
citecolor=black]{hyperref}
\usepackage{orcidlink}
\usepackage{bbding}
\def\eg{\emph{e.g.}}

\newtheoremstyle{nonindentdef}
{5pt}   
{5pt}   
{\normalfont}  
{}      
{\bfseries} 
{.}     
{ }     
{\thmname{#1}\thmnumber{ #2}\thmnote{ (#3)}} 

\theoremstyle{nonindentdef}

\newcommand{\first}[1]{\textcolor{red}{\textbf{#1}}}
\newcommand{\second}[1]{\textcolor{blue}{\textbf{#1}}}

\begin{document}

\title{LSP-ST: Ladder Shape-Biased Side-Tuning for \\Robust Infrared Small Target Detection}

\author{Guoyi~Zhang~\orcidlink{0009-0004-2931-6698}, Siyang~Chen~\orcidlink{0009-0008-9465-6647}, Guangsheng~Xu~\orcidlink{0009-0006-4235-9398}, Han~Wang~\orcidlink{0009-0008-2794-2520
	},~Donghe~Wang,~and~Xiaohu~Zhang~\orcidlink{0000-0003-4907-1451}
\thanks{Manuscript received xxx, xxx; revised xxx, xxx.}
\thanks{Equal contribution: \href{mailto:zhanggy57@mail2.sysu.edu.cn}{\textcolor{black}{Guoyi Zhang and Siyang Chen.}}}
\thanks{Corresponding authors: \textit{Han Wang, Donghe Wang and Xiaohu Zhang.}}
\thanks{Guoyi~Zhang, Siyang~Chen, Guangsheng~Xu, Han~Wang, and Xiaohu~Zhang are with the School of Aeronautics and Astronautics, Sun Yat-sen University, Shenzhen 518107, Guangdong, China.(email: \href{mailto:zhanggy57@mail2.sysu.edu.cn}{\textcolor{black}{zhanggy57}}@mail2.sysu.edu.cn;zhangxiaohu@mail.sysu.edu.cn)}
\thanks{Donghe Wang is with the Changchun Institute of Optics, Fine Mechanics and Physics, Chinese Academy of Sciences, Changchun 130033, China.}
}

\markboth{Journal of \LaTeX\ Class Files,~Vol.~14, No.~8, August~2021}%
{Guoyi Zhang \MakeLowercase{\textit{et al.}}: LSP-ST: Ladder Shape-Biased Side-Tuning for Robust Infrared Small Target Detection}


\maketitle

\begin{abstract}
Fine-tuning the Segment Anything Model (SAM) for infrared small target detection poses significant challenges due to severe domain shifts. Existing adaptation methods often incorporate handcrafted priors to bridge this gap, yet such designs limit generalization and scalability.
We identify a fundamental texture bias in foundation models, which overly depend on local texture cues for target localization. To address this, we propose Ladder Shape-Biased Side-Tuning (LSP-ST), a novel approach that introduces a shape-aware inductive bias to facilitate effective adaptation beyond texture cues. In contrast to prior work that injects explicit edge or contour features, LSP-ST models shape as a global structural prior, integrating both boundaries and internal layouts. We design a Shape-Enhanced Large-Kernel Attention Module to hierarchically and implicitly capture structural information in a fully differentiable manner, without task-specific handcrafted guidance. A theoretical analysis grounded in matched filtering and backpropagation reveals the mechanism by which the proposed attention improves structure-aware learning.
With only 4.72M learnable parameters, LSP-ST achieves state-of-the-art performance on multiple infrared small target detection benchmarks. Furthermore, its strong generalization is validated across tasks such as mirror detection, shadow detection, and camouflaged object detection, while maintaining stable performance on texture-driven tasks like salient object detection, demonstrating that the introduced shape bias complements rather than competes with texture-based reasoning.
\end{abstract}


\section{Introduction}
\IEEEPARstart{A}{lthough} fine-tuning vision foundation models \cite{SAM,SAM2,11053233}, such as the Segment Anything Model (SAM) \cite{SAM}, has shown strong performance across various downstream tasks \cite{liu2023explicit_CVPR,xiong2024sam2,10729707}, their effectiveness on infrared small target detection remains limited \cite{IRSAM}. This is largely due to the significant domain discrepancy between pretraining datasets and infrared imagery \cite{11082481}, which typically exhibits strong noise \cite{liu2023infrared}, weak texture \cite{CSRNet}, low-contrast targets \cite{DNANet}, and a severe imbalance between foreground and background information \cite{UIUNet}.

Existing methods \cite{IRSAM,11172325} typically introduce handcrafted priors during fine-tuning or incorporate scene-level priors from language models \cite{zhang2025saist} to improve infrared small target detection. However, handcrafted priors often suffer from poor generalizability \cite{11174084} and may hinder the model's end-to-end learning capacity \cite{SRNet}, limiting their applicability across tasks. Meanwhile, incorporating language models leads to a multi-modal learning paradigm \cite{zhang2025saist} that significantly increases computational overhead. Moreover, scene descriptions required by such methods are often unavailable in real-world settings, further undermining their practicality. This naturally raises a question:

\begin{tcolorbox}[
	colframe=gray!60,   
	colback=gray!5,      
	boxrule=0.8pt,         
	width=\linewidth,
	sharp corners,
	left=4pt,             
	right=4pt,
	top=4pt,
	bottom=4pt,
	fontupper=\itshape\bfseries\small 
	]
	Is it possible to achieve adaptation for infrared small target detection without compromising generalization and scalability?
\end{tcolorbox}

\begin{figure*}[!t]
	\centering
	\includegraphics[width=1\linewidth]{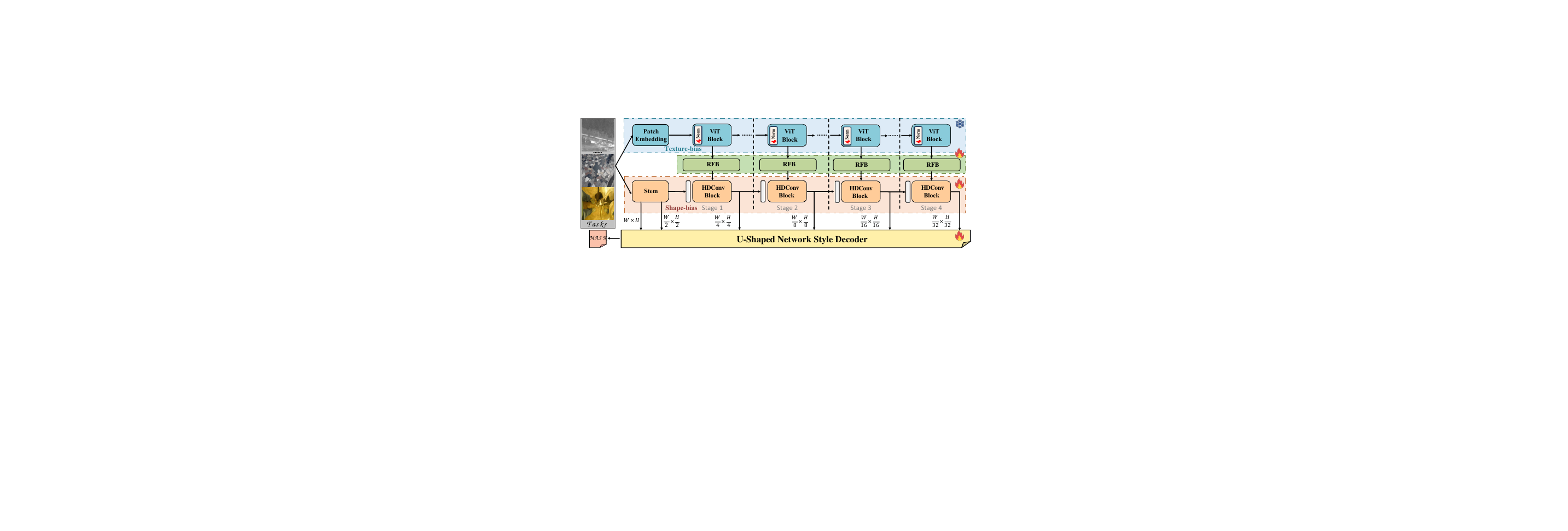}
	\caption{The overall architecture of the proposed Ladder Shape-Biased Side-Tuning (LSP-ST) framework. A dual-branch structure is constructed on top of the frozen encoder \cite{SAM2} except for the normalization layers, where the side-tuning branch introduces shape-biased representations to complement the texture-biased features from the backbone. Unidirectional connections and channel reduction are employed to achieve memory- and parameter-efficient training \cite{yin2024parameter}. Following the baseline SAM2-UNet \cite{xiong2024sam2}, we replace the original SAM2 decoder with a UNet-style decoder \cite{archit2025segment} for dense prediction.}
	\label{fig:OverallofDDNet}
\end{figure*}

In this paper, we propose Ladder Shape-Biased Side-Tuning (LSP-ST) to address the limited adaptability of vision foundation models in infrared small target detection. We observe that these models exhibit a strong texture bias~\cite{zhou2024darksam}, which impairs performance when texture cues are weak or unreliable. To mitigate this, LSP-ST introduces a shape-aware inductive bias that guides adaptation beyond texture.
As illustrated in Fig.~\ref{fig:OverallofDDNet}, the proposed architecture consists of two branches. One branch leverages a frozen ViT backbone~\cite{SAM2}, where only normalization layers are updated during tuning. The other branch is built upon the proposed HDConv blocks, which are designed to capture shape-biased representations. These two branches are connected via unidirectional lateral links to enable memory-efficient adaptation~\cite{yin2024parameter}. Additionally, a lightweight RFB module~\cite{xiong2024sam2} is used to compress feature maps and further reduce the number of tunable parameters. Finally, following our baseline setup~\cite{xiong2024sam2}, we replace the original SAM2 decoder with a U-Net-style decoder, which has been shown to be more effective for dense prediction tasks \cite{archit2025segment}.

Unlike existing task-specific methods~\cite{ISNet, SRNet, CSRNet, 10949663} that enhance shape awareness by explicitly modeling edge information, infrared imagery poses significant challenges for reliable edge extraction due to strong noise \cite{4237211}, blurred boundaries between targets and backgrounds under low signal-to-noise conditions \cite{10949663}, and imprecise annotations of small targets that introduce noise into edge supervision and limit deep model learning capacity \cite{11080263}. These practical difficulties undermine the effectiveness of explicit edge modeling. Moreover, we observe that shape and edge are fundamentally distinct: shape encompasses not only object boundaries but also the internal spatial structure of the target, both of which have been demonstrated as valuable cues for detection in classical vision studies \cite{9217948,10858405}. Motivated by these insights, we propose a shape-bias enhanced large-kernel attention module that implicitly and hierarchically captures structural information without relying on handcrafted priors or explicit edge annotations. Furthermore, we provide a rigorous theoretical analysis based on matched filtering and backpropagation, which elucidates the effectiveness of our attention mechanism.

We summarize the main contributions in this paper as follows:
\begin{itemize}
	\item We propose a novel Ladder Shape-Biased Side-Tuning (LSP-ST) framework to enhance infrared small target detection. The core insight is that the inherent texture bias in vision foundation models limits their performance on texture-insensitive tasks.
	\item Considering that shape and edge are not equivalent, we design a shape enhanced large-kernel attention module to implicitly learn object shape information, rather than relying on unstable edge cues as in existing methods.
	\item We conduct a theoretical analysis grounded in matched filtering and backpropagation to demonstrate the effectiveness of the proposed attention mechanism. (Section \ref{Sec:The})
	\item Extensive experiments across diverse downstream tasks validate the effectiveness and strong generalization capability of our method.
\end{itemize}

\begin{figure}
	\centering
	\includegraphics[width=1\linewidth]{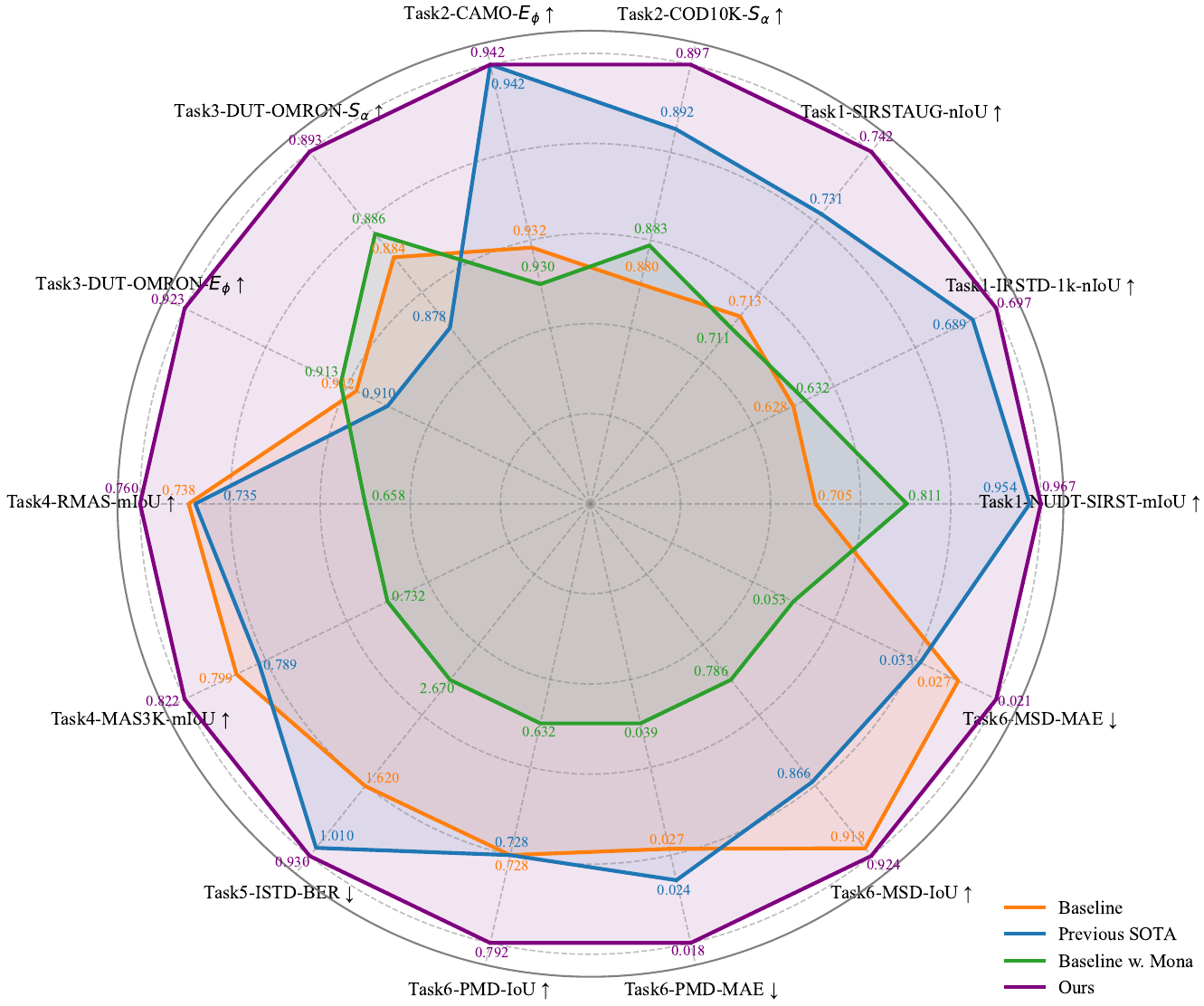}
	\caption{A radar chart is constructed based on partial results from six downstream tasks, where the baseline is SAM2-UNet \cite{xiong2024sam2}. The Previous SOTA represents the best-performing algorithms for each downstream task, excluding the baseline, including CSFwinformer (TIP'24), IRSAM (ECCV'24), SAM-SPL (TGRS'25), SAIST (CVPR'25), Dual-SAM (CVPR'24), Rmlanet (TCSVT'23), HGINet (TIP'24), UGDNet (TMM'25), AdaptCOD (IJCV'25), CamoDiffusion (TPAMI'25), and MDSAM (MM'24). It can be observed that the proposed method consistently achieves superior overall performance and outperforms the strong baseline across various tasks. In particular, conventional fine-tuning (\eg, Mona (CVPR'25) \cite{yin20255}) tends to degrade performance, whereas our approach mitigates this issue. Notably, it delivers substantial improvements on texture-insensitive tasks, while maintaining competitive results on texture-dependent ones, thereby demonstrating its strong generalization capability.}
	\label{fig:OverallofLadder}
\end{figure}
\vspace{-4pt} 
The remainder of this paper is organized as follows: Section~\ref{Section:Related_Work} briefly reviews related work. Section~\ref{Section:ProposedMethod} presents the proposed method in detail. Experimental results and analysis are provided in Section~\ref{Section:Experiment}. Finally, Section~\ref{Section:Conclusion} concludes the paper. \textbf{As shown in Fig. \ref{fig:OverallofLadder}, due to the numerous tasks, datasets, and comparison methods, including all results here would reduce readability. Detailed results for other downstream tasks (e.g., mirror, shadow, camouflaged, and salient object detection) are provided in the \underline{supplementary material}.}

\section{Related Work}\label{Section:Related_Work}
\subsection{Infrared Small Target Detection}
Deep learning-based approaches have gradually become the mainstream for infrared small target detection \cite{MSHNet}, attracting significant attention from the computer vision research community. However, existing methods are predominantly task-specific model designs \cite{DNANet,ISNet,UIUNet}. Although some studies \cite{IRSAM,zhang2025saist,11172325} have attempted to unleash the potential of visual foundation models for infrared small target detection, most of them rely on incorporating task-specific handcrafted priors during fine-tuning, such as wavelet-domain priors \cite{IRSAM} or the generation of textual scene descriptions \cite{zhang2025saist}. These methods, however, compromise the generalizability of the fine-tuning process across different tasks.

In contrast to the above methods, our key observation lies in the inherent texture bias of visual foundation models \cite{zhou2024darksam}, which hinders their adaptability to infrared small target detection. To address this issue, we propose Ladder Shape-Biased Side-Tuning, a method that does not rely on task-specific handcrafted priors. We further demonstrate the generalization ability of the proposed approach across multiple natural image tasks.
\subsection{Parameter-Efficient Fine-Tuning}
Vision foundation models pretrained on large-scale datasets have demonstrated strong generalization and visual recognition capabilities \cite{SAM}. To harness this potential in downstream tasks without incurring the high computational cost of full fine-tuning, parameter-efficient fine-tuning (PEFT) methods have emerged as a practical alternative \cite{11080219}.
Adapter-based methods \cite{c4} and reparameterization techniques such as LoRA \cite{11080219} have gained popularity across various applications. However, their performance often degrades on dense prediction tasks, particularly when there is a significant domain gap between the pretraining and downstream data. The success of VIT-Adapter \cite{chen2022vision} has highlighted the effectiveness of side-tuning approaches in such scenarios, spurring further research into more efficient architectures. Following this, methods like LST \cite{sung2022lst} and Vit-comer \cite{xia2024vit} have been introduced to reduce the number of trainable parameters while enhancing model expressiveness.

Despite these advancements, existing PEFT approaches still struggle with complex downstream tasks. Some recent works \cite{IRSAM} attempt to mitigate this by incorporating handcrafted task-specific priors during fine-tuning. While this can improve task-specific performance, it compromises the end-to-end optimization process and often limits generalization to other tasks.
Recent studies \cite{zhou2024darksam} have also revealed that vision foundation models such as SAM \cite{SAM} exhibit a strong texture bias. Moreover, current PEFT techniques may inadvertently amplify this bias, hindering performance on texture-insensitive tasks. This limitation becomes particularly pronounced in infrared small target detection, where weak texture signals, extreme foreground-background imbalance, low signal-to-noise ratios, and complex background clutter collectively pose significant challenges.

\begin{figure}
	\centering
	\includegraphics[width=1\linewidth]{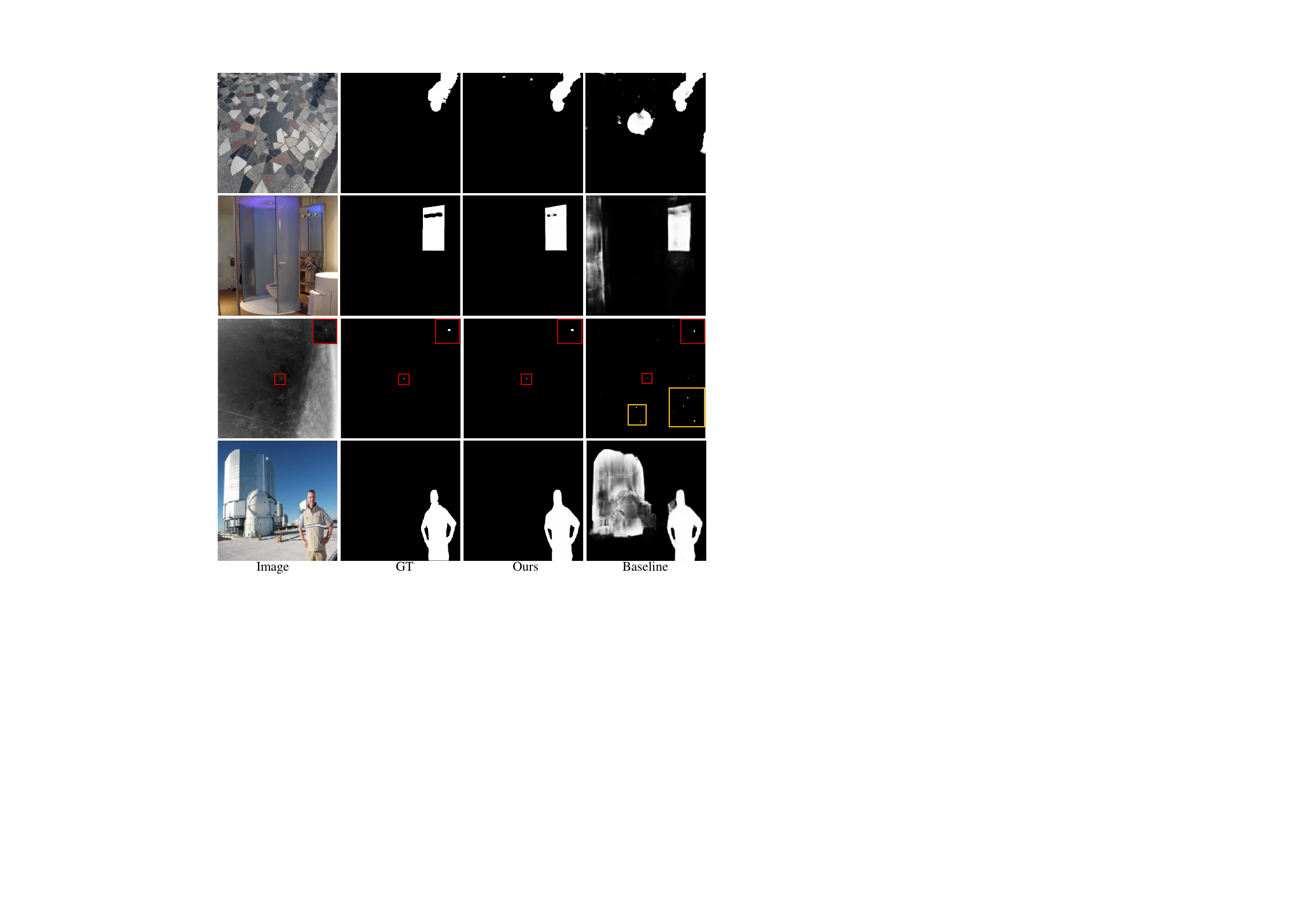}
	\caption{Comparison of visual recognition results under complex scene texture interference. It can be observed that the proposed method demonstrates greater robustness in these scenarios.}
	\label{fig:OverallofMSD}
\end{figure}
To address these issues while preserving generalization, we propose a fine-tuning strategy that introduces shape-biased representation learning. As shown in Fig. \ref{fig:OverallofMSD}, our method minimizes reliance on handcrafted priors and demonstrates consistent effectiveness across a diverse range of texture-insensitive tasks.
\subsection{Shape Bias in Representation Learning}
Shape-biased representations constitute an essential component of representation learning \cite{ISNet,SRNet,CSRNet}, and the incorporation of shape information has been shown to improve performance in various downstream tasks \cite{8603790,10844993}. While most existing methods focus on classification tasks and incorporate shape regularization through novel data augmentation strategies or customized loss functions \cite{he2023shift}, approaches in dense prediction tasks \cite{7458161} often conflate shape with contour. These methods predominantly rely on edge cues \cite{10705912} and enlarged receptive fields \cite{SRNet,CSRNet} to construct shape-biased representations, which may lead to an oversimplified understanding of the underlying shape semantics.

\begin{figure}
	\centering
	\includegraphics[width=1\linewidth]{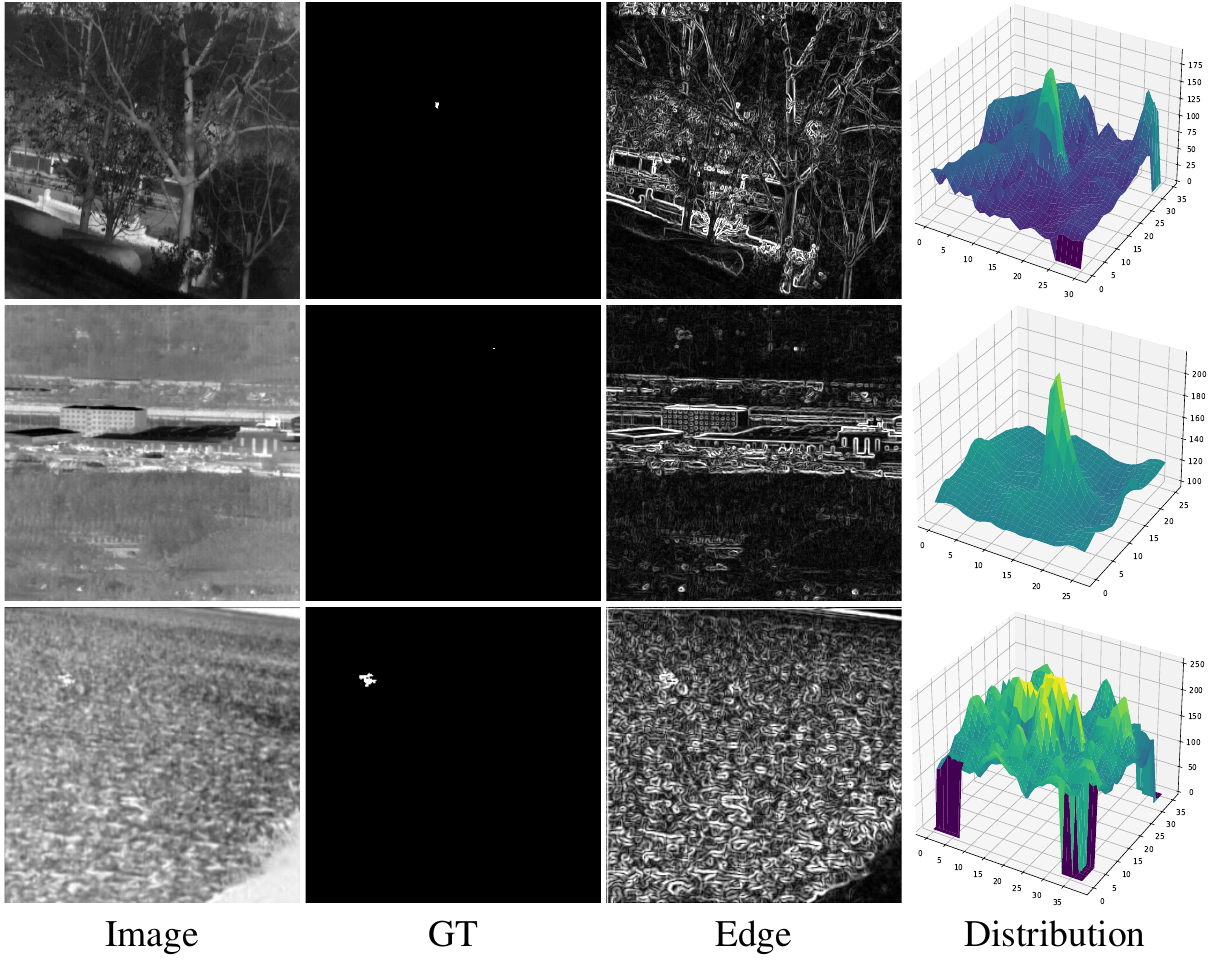}
	\caption{In complex scenes, intra-object distributions better capture target characteristics than boundary cues. Together, they jointly define the overall shape of the object.}
	\label{fig:OverallofShape}
\end{figure}

In contrast to prior work, we observe that shape and contour are not equivalent; notably, as shown in Fig. \ref{fig:OverallofShape}, shape encompasses both internal distribution characteristics and contour information.
Specifically, to enable the learning of shape-biased representations, the model must comprehensively capture information from both the target and its surrounding background, thereby facilitating the characterization of contour features. Additionally, the model should be sensitive to the internal distribution characteristics of the target and focus as much as possible on the target interior to avoid feature entanglement between the target and background. 
To this end, rather than directly designing handcrafted priors, we propose a shape-enhanced large-kernel attention mechanism and provide theoretical analysis to demonstrate its effectiveness.
\section{Proposed Method} \label{Section:ProposedMethod}
As shown in Fig.~\ref{fig:OverallofDDNet}, the ViT branch in our method is based on the Hiera-ViT backbone from SAM2 \cite{SAM2}, whose multi-scale representation capability has been widely demonstrated to be effective for dense prediction tasks \cite{chen2022vision}. Furthermore, considering that small targets require low-level semantic cues for feature refinement \cite{11156113}, we incorporate high-resolution feature maps from the stem into the decoder to enhance spatial detail recovery.  
To ensure a fair comparison and better evaluate our representational capability, we adopt the same RFB module as the baseline for channel compression and follow identical hyperparameter settings \cite{xiong2024sam2}. Feature maps from all stages are uniformly reduced to 32 channels.  
In addition, we replace the original SAM2 decoder with the standard U-Net decoder. Since our method targets frame-wise inference rather than video-based processing, all temporal modules in SAM2 are removed.

\subsection{Ladder Shape-Biased Side-Tuning}  \label{LSR-ST}
\begin{figure}
	\centering
	\includegraphics[width=1\linewidth]{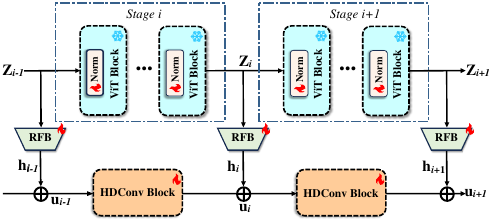}
	\caption{Detailed architecture of the proposed LSP-ST. To reduce the number of learnable parameters, for each stage, we first perform feature dimensionality reduction via the RFB module \cite{xiong2024sam2} and incorporate only a single instance of the proposed HDConv Block.}
	\label{fig:OverallofSB}
\end{figure}
The details of the proposed Ladder Shape-Biased Side-Tuning (LSP-ST) framework are illustrated in Fig.~\ref{fig:OverallofSB}. The design is both simple and effective. After fine-tuning, the learnable structures within the encoder consist of three parts: (1) Normalization Layers in the Frozen ViT Backbone. These layers are retained for fine-tuning while keeping the rest of the backbone frozen. By modulating the mean and variance of feature maps, they adjust the feature distribution and enable the backbone to capture task-specific knowledge with minimal parameter updates \cite{10.1145/3664647.3680834}. (2) RFB module. This module is used for channel compression, reducing the dimensionality of feature maps to improve parameter efficiency. This design remains consistent with our baseline \cite{xiong2024sam2}. (3) HDConv Blocks. As the core of the proposed tuning mechanism, these blocks are designed to capture shape-aware representations. By focusing on structural rather than textural information, they enhance model robustness under weak-texture conditions.

\noindent\textbf{Detail of the proposed HDConv Block.} The HDConv Block serves as the core component of the proposed LSP-ST framework and represents the primary distinction from conventional fine-tuning strategies. The entire side-tuning branch is constructed by stacking multiple HDConv Blocks, enabling the network to effectively capture the desired shape-biased representations. The HDConv Block is constructed by stacking several identical basic units. For each individual unit, given an input feature map $\mathbf{X}$, the processing pipeline proceeds as follows:
\begin{equation}
	\mathbf{X}^{\prime}=\mathbf{X}+\lambda_1\text{SPE-Attention}\left(\mathrm{GroupNorm}(\mathbf{X})\right),
\end{equation}
\begin{equation}
	\mathbf{X}^{\prime\prime}=\mathbf{X}^{\prime}+\lambda_2\text{Pre-BasicBlock}\left(\text{Pre-BasicBlock}\left(\mathbf{X}^{\prime}\right)\right).
\end{equation}
where $\mathrm{GroupNorm}(\cdot)$ denotes Group Normalization, $\lambda_1$ and $\lambda_2$ are learnable parameters in the LayerScale \cite{touvron2021going} mechanism. $\text{Pre-BasicBlock}$ denotes pre-activation residual block \cite{huang2023revisiting}, and $\text{SPE-Attention}(\cdot)$ denotes proposed shape enhanced large kernel attention (as shown in Fig. \ref{fig:Overallofattention}.).

\subsection{Shape Enhanced Large-kernel Attention Mechanism} \label{SE}
\begin{figure}
	\centering
	\includegraphics[width=1\linewidth]{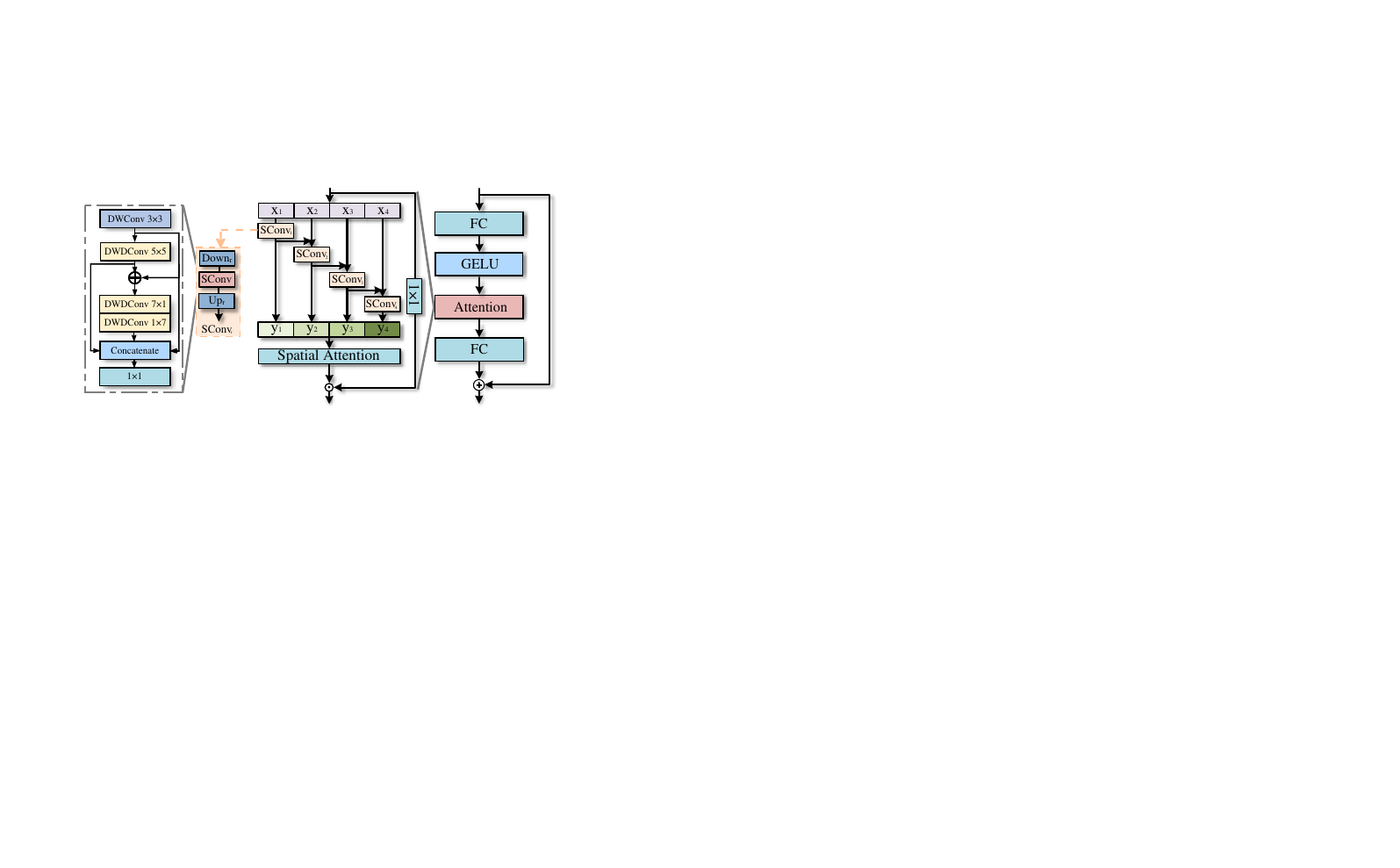}
	\caption{The overall architecture of the proposed Shape Enhanced Large-kernel Attention Mechanism is both simple and effective, and it is theoretically justified, as detailed in Section \ref{Sec:The}.}
	\label{fig:Overallofattention}
\end{figure}

Unlike prior approaches that describe object shape primarily through contour information, we argue that shape and contour are inherently different. Shape should be understood as a more holistic concept that includes both the boundary and the internal spatial distribution of a target. This distinction becomes particularly important in infrared small target detection, where annotations are often imprecise and the boundaries between foreground and background tend to be ambiguous due to noise and low contrast. In such cases, using contour information either as input or supervision can introduce considerable noise. Additionally, contour-based methods often require complex decoder-side modules to extract and refine boundary cues, which increases the overall model complexity.

\noindent\textbf{Overview of the proposed Shape-Enhanced Large-Kernel Attention.} To overcome these limitations, we introduce the Shape-Enhanced Large-Kernel Attention (SELKA) module, as shown in Fig.~\ref{fig:Overallofattention}. SELKA is designed to capture both the internal distribution patterns and boundary characteristics of a target, enabling more effective learning of shape-biased representations. Given an input feature map $\mathbf{X}$, the processing pipeline proceeds as follows:
\begin{equation}
\hat{\mathbf{X}} = \text{FC}(\text{Attention}(\text{GELU}(\text{FC}(\mathbf{X})))) + \mathbf{X},
\end{equation}
where, FC refers to a 1$\times$1 convolution, and GELU denotes the activation function. Attention represents the core component, for an input feature map \( \mathbf{X} \in \mathbb{R}^{W \times H \times C} \), the attention mechanism computes the similarity score matrix \( \mathbf{A} \in \mathbb{R}^{W \times H \times C} \) and the value matrix \( \mathbf{V} \in \mathbb{R}^{W \times H \times C} \). Then, it uses the Hadamard product to calculate the output \( \mathbf{Z} \in \mathbb{R}^{W \times H \times C} \). The entire process is as follows:
\begin{equation}
	\mathbf{Z} = \text{SiLU}(\mathbf{A})\odot\text{SiLU}(\mathbf{V}), \label{Eq:AV}
\end{equation}
where $\text{SiLU}(\cdot)$ is sigmoid linear unit and the value matrix \( \mathbf{V} \) is computed using a simple \( 1 \times 1 \) convolution. 

\noindent\textbf{Detail of the computation of the similarity score matrix $\mathbf{A}$.} The computation of the similarity score matrix is key to the attention mechanism. First, we adopt large-kernel convolutions as the fundamental building block. This choice is motivated by the need to capture both internal and external features of the target in order to accurately delineate its contour. Achieving this requires a large effective receptive field. At the same time, due to the localized nature of small targets, the model must possess strong inductive biases to effectively capture such features. Given an input feature map $\mathbf{X}$, the processing procedure of the proposed SConv large-kernel convolution operator is as follows:
\begin{equation}
	\mathbf{X}_{D3} = \text{DWConv}_{3\times3}(\mathbf{X}), 
\end{equation}
\begin{equation}
	\mathbf{X}_{D5} = \text{DWConv}_{5\times5}(\mathbf{X}_{D3}), 
\end{equation}
\begin{equation}
	\mathbf{X}_{D7} = \text{DWConv}_{7\times1}(\text{DWConv}_{1\times7}(\mathbf{X}_{D3} + \mathbf{X}_{D5}))), 
\end{equation}
\begin{equation}
	\hat{\mathbf{X}} = \text{Conv}_{1\times1}(\text{Concatenate}(\mathbf{X}_{D3}, \mathbf{X}_{D5}, \mathbf{X}_{D7}).
\end{equation}
Specifically, the 3$\times$3 and 5$\times$5 depthwise convolutions capture the internal distribution of features, while the decomposed 1$\times$7 and 7$\times$1 strip-shaped convolutions are sensitive to boundaries. These feature maps are concatenated to integrate different aspects of the target. Afterwards, a 1$\times$1 convolution is applied for dimensionality reduction and channel-wise information interaction.

To further account for the multi-scale nature of the targets, we employ a multi-branch architecture where each branch applies different downsampling rates to increase the effective receptive field. The features from these branches are then fused through interaction. Given an input feature map $\mathbf{X}$, the processing procedure is as follows:
\begin{equation}
	\left\{\mathbf{U}_1, ..., \mathbf{U}_r\right\} = \text{Split}(\mathbf{X}),
\end{equation}
\begin{equation}\label{Eq:Cass_SS}
	\hat{\mathbf{U}}_j=\begin{cases}
		\uparrow_p\text{ (SConv}(\downarrow_{\frac p{2^j}}(\mathbf{U}_j))),j = 1. \\
		\uparrow_p\text{ (SConv}(\downarrow_{\frac p{2^j}}(\mathbf{U}_j + \hat{\mathbf{U}}_{j-1}))),j \textgreater 1.
	\end{cases} 
\end{equation}
where $\uparrow_{p} (\cdot)$ represents upsampling features at a specific level to the original resolution
$p$  via nearest interpolation for fast implementation and $\downarrow\frac p{2^j} (\cdot)$ denotes maxpooling the input features to the size of $\frac p{2^j}$. 

Finally, a simple spatial attention mechanism is employed to selectively fuse features from multiple branches. Given a set of feature maps $\left\{\hat{\mathbf{U}}_1, ..., \hat{\mathbf{U}}_r\right\}$, the processing procedure is as follows:
\begin{equation}
	\hat{\mathbf{U}}_1^{avg} = \text{Pooling}_{avg}(\hat{\mathbf{U}}_1), \ \hat{\mathbf{U}}_1^{max} = \text{Pooling}_{max}(\hat{\mathbf{U}}_1),
\end{equation}
\begin{equation}
	\mathbf{W} = \text{Concatenate}(\left\{\hat{\mathbf{U}}_1^{avg}, ..., \hat{\mathbf{U}}_r^{max}\right\}),
\end{equation}
\begin{equation}
	\left\{\hat{\mathbf{W}}_1, ..., \hat{\mathbf{W}}_r\right\} = \text{Split}(\text{Conv}_{7\times7}(\mathbf{W})),
\end{equation}
\begin{equation}
	\mathbf{A} = \text{Concatenate}(\left\{\hat{\mathbf{U}}_1 \cdot  \hat{\mathbf{W}}_1, ..., \hat{\mathbf{U}}_r \cdot  \hat{\mathbf{W}}_r\right\}),
\end{equation}
where the feature map $\mathbf{A}$ corresponds to the similarity score matrix. 

\subsection{Theoretical Analysis of the Effectiveness of the Proposed Attention Mechanism} \label{Sec:The}
\begin{table*}[ht]
	\centering
	\caption{Comparison of Theoretical Analysis Paradigms}
	\label{tab:paradigm_compare}
	\begin{tabular}{l|c|c|c|c|c|c|c}
		\toprule
		\textbf{Method} &
		\textbf{Physical Prior} &
		\textbf{Theory} &
		\textbf{Design Guidance} &
		\textbf{Perceptual Bias} &
		\textbf{Adaptivity} &
		\textbf{Interpretability} &
		\textbf{Model Stability} \\
		\midrule
		Forward Design\dag &
		\Checkmark\kern-1.2ex\raisebox{1ex}{\rotatebox[origin=c]{125}{\textbf{--}}} &
		\Checkmark\kern-1.2ex\raisebox{1ex}{\rotatebox[origin=c]{125}{\textbf{--}}} &
		\Checkmark\kern-1.2ex\raisebox{1ex}{\rotatebox[origin=c]{125}{\textbf{--}}} &
		\XSolidBrush &
		\XSolidBrush &
		\Checkmark\kern-1.2ex\raisebox{1ex}{\rotatebox[origin=c]{125}{\textbf{--}}} &
		\XSolidBrush \\
		Attribution Analysis\ddag &
		\XSolidBrush &
		 \CheckmarkBold &
		\XSolidBrush &
		\XSolidBrush &
		\Checkmark\kern-1.2ex\raisebox{1ex}{\rotatebox[origin=c]{125}{\textbf{--}}}&
		\Checkmark\kern-1.2ex\raisebox{1ex}{\rotatebox[origin=c]{125}{\textbf{--}}} &
		\XSolidBrush \\ \rowcolor{gray!20}
		\textbf{Ours}¶ &
		\CheckmarkBold &
		\CheckmarkBold &
		\CheckmarkBold &
		\CheckmarkBold &
		\CheckmarkBold &
		\CheckmarkBold &
		\CheckmarkBold \\
		\bottomrule
		\multicolumn{8}{l}{\footnotesize{(\dag) Due to the black-box nature of neural networks, forward design (\eg, RPCANet \cite{RPCANet}) often lacks rigorous enforcement of physical priors.}}\\
		\multicolumn{8}{l}{\footnotesize{(\ddag) Attribution analysis (\eg, CAM \cite{zhou2016learning}) results are sensitive to sample selection and often do not provide direct guidance for model design.}} \\
		\multicolumn{8}{l}{\footnotesize{(¶) This work uses backpropagation to measure perceptual biases in fixed-parameter neural networks and align them with target characteristics.}}
	\end{tabular}
\end{table*}
To substantiate the capability of the proposed attention mechanism in capturing target shape bias, particularly contours and internal distribution patterns, we present a theoretical analysis grounded in signal detection theory and neural sensitivity modeling, the strengths of our theoretical analysis are presented in Tab. \ref{tab:paradigm_compare}.
We begin by examining the intrinsic distribution characteristics of the target\footnote{For clarity, we focus on concrete infrared small targets. However, with the introduction of RFLA \cite{xu2022rfla} or SAFit \cite{10914515}, the analysis can be readily extended to general small object detection (even morphologically extended small objects) with minimal changes.}. Next, we analyze the necessary conditions that a neural network must satisfy to be sensitive to such internal variations. Finally, we show that our design meets these conditions both theoretically and empirically. Moreover, by incorporating the concept of heat conduction \cite{wang2025building}, we extend the effectiveness of our method to general natural scene visual tasks.

\subsubsection{Intrinsic Distribution Characteristics of Targets} 
Robust detection of infrared small targets requires distinguishing feature correlations between true targets and background-induced false alarms~\cite{liu2023infrared}. To achieve this, it is crucial to introduce both target-relevant and false-alarm-related information into deeper network layers \cite{UIUNet}. However, due to the weak intensity and limited spatial extent of both infrared targets and false alarms, these signals are highly susceptible to degradation during deep feature propagation \cite{DNANet}.
The imaging process for both types of signals can be modeled as a convolution of the intrinsic radiation distribution with the point spread function (PSF) \cite{RDIAN}, which accounts for atmospheric scattering and diffraction effects introduced by the imaging system. For small and dim structures, this convolution is predominantly governed by the PSF \cite{Dai_2021_WACV}, which typically resembles a Gaussian distribution under diffraction-limited conditions~\cite{zhang2024dtnet}.

\subsubsection{Conditions a Neural Network Should Satisfy} 
Since large targets have already been effectively addressed in practical scenarios \cite{11172325}, our analysis focuses on point-like small targets. These targets are particularly prone to information loss during feature propagation \cite{11146868}. The same analysis also applies to slightly extended small targets exhibiting basic shape characteristics \cite{11156113}, as the PSF remains a dominant factor compared to shape variability. Specifically, the observed target signature $I_T(x, y)$ is governed by the convolution of a point-like target with the PSF, expressed as \cite{zhang2024dtnet}
\begin{equation}
	I_T(x, y) = A \cdot \mathrm{PSF}(x, y),
	\label{eq:psf_target}
\end{equation}
where $A$ denotes target intensity. The detection task can be formulated as a classical hypothesis testing problem, in which the observed image contains either only noise (null hypothesis) or a known deterministic signal embedded in noise (alternative hypothesis). If a linear filter $H(x, y)$ is applied to the image, the resulting output response is:
\begin{equation}
		R = A \cdot \iint \mathrm{PSF}(x, y) \cdot H(x, y) dxdy.
\end{equation}
To maximize the response $R$ under fixed energy of $H(x, y)$, the Cauchy-Schwarz inequality gives
\begin{equation}
	R \leq A \cdot \left( \iint \mathrm{PSF}^2(x, y)dxdy \right)^{\frac{1}{2}} \cdot \left( \iint H^2(x, y)dxdy \right)^{\frac{1}{2}}
\end{equation}
with equality if and only if $H(x,y)=c \cdot \mathrm{PSF}(x,y)$ for some constant $c$. Therefore, the optimal linear filter that maximizes detection response under PSF-dominant conditions is the matched filter:
\begin{equation}
	H^*(x, y) = \mathrm{PSF}(x, y).
\end{equation}
Therefore, in the classical detection framework, the PSF-shaped matched filter maximizes the SNR and serves as an optimal response template. This implies that a neural network designed for such detection tasks should implicitly learn spatial sensitivity patterns\footnote{Such spatial sensitivity patterns are closely related to the notion of context \cite{li2024lsknet}, as they define the boundary of contextual awareness. The conveyed contextual information is subject to modulation by the $\phi(I)$ in Eq. (\ref{Eq:NTK}).} that align with $H^*(x, y)$.
While $H^*(x, y)$ arises from linear analysis, it offers valuable guidance regarding the desired spatial response of an effective model. Inspired by the theory of neural tangent kernels (NTK)~\cite{jacot2018neural}, we approximate the network's behavior as
\begin{equation}
	f(I(x,y)) \approx \text{Agg}(\phi(I(x,y)), S(x, y)), \label{Eq:NTK}
\end{equation}
where $\text{Agg}(\cdot)$ models the interaction between nonlinear transformations and spatially-aware representations. Although this is not a direct consequence of NTK theory, the decomposition is motivated by its interpretation of neural networks as structured kernel machines~\cite{jacot2018neural}.
Here, $\phi(I)$ denotes the nonlinear transformation applied to the input $I$, and $S(x, y)$ represents the spatial sensitivity pattern. Ideally, $S(x, y)$ should resemble the matched filter $H^*(x, y)$. In the context of small infrared targets~\cite{li2023ilnet}, the contribution from $\phi(I)$ is often limited\footnote{Nonlinear representational capacity $\phi(I)$ may be analyzed via feature subspace dimensionality \cite{ma2024rewrite}, notingthat the injection of foundation model features expands the initial subspace \cite{10.1145/3664647.3680834}.}~\cite{RDIAN}, making $S(x, y)$ the principal component determining detection performance. 
To ensure that the model's decision-making aligns with the behavior of the ideal filter, we analyze gradient dynamics during
backpropagation\footnote{A fixed-parameter neural network functions as a nonlinear filter, whose intrinsic feature preferences can be revealed through gradient backpropagation on inputs drawn from different distributions \cite{zhou2016learning}.}.

To quantify this alignment, we define two metrics. First, the gradient variation under a small input perturbation $\Delta I$ (e.g., spatial shift or distributional change) is given by:
\begin{equation}
	\delta_{\text{grad}} = \left\| \frac{\partial f(I + \Delta I)}{\partial I} - \frac{\partial f(I)}{\partial I} \right\|_2,
\end{equation}
which reflects the stability of the network’s gradient with respect to input variation. Second, the filter alignment score is defined as:
\begin{equation}
	\mathcal{A}_{\text{align}} = \frac{ \left\langle \frac{\partial f(I)}{\partial I}, H^* \right\rangle }{ \left\| \frac{\partial f(I)}{\partial I} \right\|_2 \cdot \left\| H^* \right\|_2 } = \frac{\sum_{x, y} \frac{\partial f(I)}{\partial I(x, y)} \cdot H^*(x, y)}{ \left\| \frac{\partial f(I)}{\partial I} \right\|_2 \cdot \left\| H^* \right\|_2}.
\end{equation}
A well-structured network should exhibit high alignment \( \mathcal{A}_{\text{align}} \) and low gradient variation \( \delta_{\text{grad}} \) under mild perturbations. 
Although this gradient-based analysis provides intuitive insight, it is challenging to directly translate it into architectural design principles. Notably, this analysis closely aligns with the concept of the \textit{effective receptive field} (ERF) \cite{luo2016understanding}. Specifically, the ERF characterizes the actual spatial influence of input pixels on the output activations in trained networks. It functions as a data-driven approximation of the network’s sensitivity map. In fact, \( \mathcal{A}_{\text{align}} \) defines the alignment between the ERF and $H^*(x,y)$.
We thus formulate the structural alignment requirement as a shape-matching condition\footnote{It is worth noting that, as a by-product of our analysis, we challenge the common tendency in many studies \cite{ding2024unireplknet,DBLP:conf/iclr/LiuCCCXWKPMW23,finder2025wavelet} to blindly pursue a larger effective receptive field (\eg, global \cite{liu2023infrared}). Our findings suggest that an oversized ERF is not inherently beneficial; instead, it should be aligned with the target’s intrinsic characteristics in both size and shape. This observation is consistent with empirical findings reported in prior works~\cite{11029687,11069297,li2024lsknet}.}: 
\begin{equation}
	\min_{\text{ERF}}\mathcal{L}_{\text{ERF}}=\min_{\text{ERF}} \|\text{ERF}(x, y) - H^*(x, y) \|^2_2. \label{Eq:Match}
\end{equation}
An effective architecture should therefore be intrinsically sensitive to spatial structures induced by the PSF.

\subsubsection{Architectural Correspondence to the Theoretical Conditions}
\begin{figure}
	\centering
	\includegraphics[width=1\linewidth]{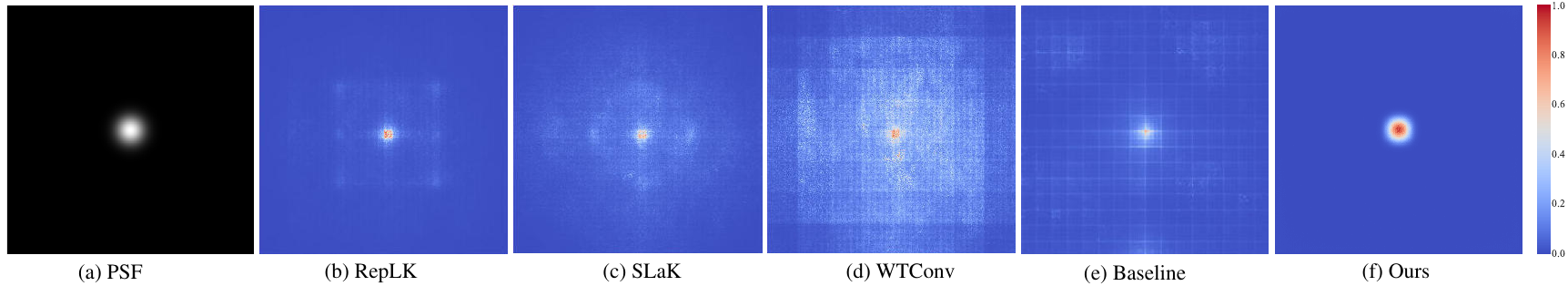}
	\caption{Comparison of effective receptive field (ERF) shapes for different large-kernel convolutions (\eg, RepLK (CVPR'22) \cite{ding2022scaling}, SLaK (ICLR'23) \cite{DBLP:conf/iclr/LiuCCCXWKPMW23}, and WTConv (ECCV'24) \cite{finder2025wavelet}). The ERF of the proposed SConv module exhibits a Gaussian-like distribution, closely approximating the PSF and enhancing sensitivity to target structures dominated by optical blurring. This enables SConv to be more sensitive to the internal distribution of small targets significantly affected by the PSF, thereby enhancing the capture of shape bias.}
	\label{fig:OverallofERF}
\end{figure}
Having established the theoretical conditions under which the ERF must align with PSF-induced patterns, we now show that our architecture meets these alignment constraints. We first analyze how the proposed SConv module meets the alignment conditions at the operator level. Based on this, we then illustrate how the overall attention mechanism achieves structural correspondence by aggregating multiple SConv-based branches.

\noindent\textbf{The proposed SConv.} Empirically, as reported in \cite{luo2016understanding}, convolutions with small kernels (kernel size $K<7$) yield an ERF that approximates a Gaussian distribution. In the proposed SConv module, features are aggregated through residual connections, effectively summing multiple Gaussian-like responses with varying variances, resulting in an overall suitable Gaussian-shaped ERF \footnote{Short skip connections enable the model to adaptively bypass certain layers during training \cite{ding2022scaling}, thereby regulating the effective depth $N$ in Eq. (\ref{Eq:ERFS}). This facilitates the formation of a receptive field with a suitable size \cite{luo2016understanding} that aligns with the ideal $H^*(x,y)$. For further details, please refer to the short-path theory \cite{veit2016residual}, where the residual structure is interpreted as an ensemble of multiple paths of varying lengths.}. However, small kernels impose limited ERF sizes \cite{ding2024unireplknet}, which hinders capturing the structural characteristics of the target and achieving shape bias \cite{CSRNet}. When $K \geq 7$, the ERF often deviates from Gaussian form \cite{DBLP:journals/corr/abs-2207-03620}. Moreover, the ERF size approximately satisfies \cite{luo2016understanding}
\begin{equation}
	Size_{\text{ERF}} \propto K\sqrt{N}, \label{Eq:ERFS}
\end{equation}
where $N$ is the effective network depth. Consequently, large kernels tend to dominate the ERF in SConv, potentially leading to spatial responses that diverge from the desired Gaussian form.
To mitigate this issue, large square kernels tend to cause excessive feature mixing between small targets and background within the receptive field~\cite{MRF3Net}. Decomposing them into strip-shaped convolutions reduces this effect by selectively capturing directional boundary information~\cite{guo2022segnext}, thereby enhancing shape sensitivity~\cite{lau2024large}. Frequency analysis further reveals that strip convolutions act as directional band-pass filters~\cite{8936525}, emphasizing anisotropic structures such as edges and ridges essential for preserving shape cues. Moreover, kernel decomposition helps maintain the ERF's Gaussian-like profile by mitigating the dominance of large kernels. As shown in Fig.~\ref{fig:OverallofERF}, SConv hierarchically aggregates both small- and large-kernel features, reinforcing the overall Gaussian property of the ERF (obtained through empirical measurement).

\noindent\textbf{The proposed Attention.} While the ERF induced by a single SConv operator approximates a Gaussian shape, using a fixed ERF distribution remains suboptimal. Even with a fixed infrared imaging sensor, PSF variations across scenes limit the effectiveness of a single fixed ERF shape, as defined in Eq.~(\ref{Eq:Match}), since a single canonical ERF cannot accommodate the full range of PSF-induced distortions. According to Fraunhofer diffraction theory~\cite{zhang2024dtnet}, the PSF resulting from diffraction can be modeled as the squared magnitude of a first-order Bessel function:
\begin{equation}
	O(\lambda, r) = \left[ \frac{2 J_1\left( \frac{\pi D r}{\lambda f} \right)}{ \frac{\pi D r}{\lambda f} } \right]^2,
\end{equation}
where $J_1(\cdot)$ is the first-order Bessel function, $D$ is the aperture diameter, $\lambda$ the wavelength, $f$ the focal length, and $r$ the radial distance from the optical axis. Fortunately, for any fixed wavelength $\lambda$, the corresponding PSF can be well approximated by a Gaussian distribution with suitable parameters. Leveraging this approximation, the proposed attention mechanism can be interpreted as a parametric ERF generator, dynamically assembling ERFs with varying scales and variances to approximate different PSF profiles. Based on the ERF formulation in Eq.~(\ref{Eq:ERFS}), the size of the ERF can be modulated by adjusting interaction patterns, equivalent network depth, or convolution kernel sizes. 

As shown in Fig.~\ref{fig:Overallofattention}, all attention branches employ the same SConv operator, producing Gaussian-shaped ERFs. Spatial scaling operations (e.g., upsampling or downsampling) stretch the ERF but do not alter its shape, as Gaussian functions are scale-invariant.  Formally, given an original ERF distribution $\text{ERF}_{\text{orig}} (i, j)$, the ERF after spatial scaling by factor $s$ (downsampling if $s > 1$, upsampling if $s < 1$) is
\begin{equation}
	\text{ERF}_{s}(i, j) = \frac{1}{s^2} \cdot \text{ERF}_{\text{orig}}\left( \frac{i}{s}, \frac{j}{s} \right).
\end{equation}
This property follows from scale-space theory \cite{lafferty2005diffusion}, which identifies the Gaussian kernel as the only function preserving structural consistency under scaling transformations. Consequently, aggregating responses from multiple attention branches corresponds to summing several Gaussian distributions at different scales, yielding an overall ERF that retains its Gaussian shape. The lightweight spatial attention module embedded within our attention mechanism adaptively selects the ERF configurations that best align with the scene-specific PSF structures. By maintaining a Gaussian-like ERF across scales, the model achieves both structural sensitivity and SNR-optimality in detecting small targets dominated by optical blur.
\subsubsection{Extended Analysis: From Specific to General} 
Theoretical analysis confirms the proposed method’s effectiveness for infrared small target detection. Unlike prior work \cite{IRSAM,zhang2025saist}, it avoids fine-tuning that harms generalization and scalability, achieving a dynamically scale-adaptive Gaussian receptive field fundamental to shape analysis in diverse scenarios. To elucidate this point, it is necessary to introduce the lens of heat diffusion\footnote{The heat equation has long been used in computer vision as a theoretical tool for modeling shape evolution and structural regularity~\cite{1344036,7812756}, which aligns well with our goal of capturing shape bias.}~\cite{1344036}. In continuous space, the process of isotropic heat conduction is governed by the heat equation~\cite{wang2025building}:
\begin{equation}
	\frac{\partial u}{\partial t} = \alpha \nabla^2 u,
\end{equation}
where \( u(x, y, t) \) denotes the temperature distribution and \( \alpha \) is the diffusion coefficient. The fundamental solution to this PDE with a point source initial condition is the Gaussian heat kernel \cite{lafferty2005diffusion}:
\begin{equation}
	G(x, y; t) = \frac{1}{4\pi \alpha t} \exp\left(-\frac{x^2 + y^2}{4\alpha t}\right). \label{Eq:dif}
\end{equation}
This kernel describes how information diffuses over space with time \( t \), where the variance \( \sigma^2 = 2\alpha t \) controls the diffusion scale \footnote{While anisotropic Gaussian receptive fields can improve task-specific performance, their sensitivity to shape deviations may lead to performance degradation. In contrast, isotropic Gaussians with appropriate scale provide a more robust and generalizable alternative.}
. The kernel size varies across different stages and also depends on the specific characteristics of each image.

Our method, by constructing the ERF as a Gaussian with dynamic scale, inherently aligns with Eq. (\ref{Eq:dif}).
Therefore, the proposed attention mechanism possesses the ability to capture shape information similar to heat diffusion. Since shape analysis is crucial for many natural scene tasks\footnote{Interestingly, similar characteristics have also been observed in the human visual perception system \cite{luo2016understanding}, which also demonstrates strong shape modeling and perception capabilities \cite{he2023shift}.}~\cite{4032826,1632205}, this enables our method to adapt to infrared small target detection without sacrificing generalization. The experimental results of the infrared small target detection task are presented in Section \ref{Section:Experiment}, while the effectiveness of the Gaussian-shaped effective receptive field is validated in Section \ref{Sec:Gau}. Further experimental validation of this generalization capability is provided in the \underline{\textbf{supplementary material}}.
\subsection{Loss Function}
Consistent with previous infrared small target detection works, we adopt the same loss function (Soft-IoU Loss) to address the severe imbalance between target foreground and background.
\begin{equation}
	\mathcal{L} = 1 - \frac{\sum_{i,j}\mathbf{X}_{i,j}\cdot\mathbf{Y}_{i,j}}{\sum_{i,j}\mathbf{X}_{i,j}+\mathbf{Y}_{i,j}-\mathbf{X}_{i,j}\cdot\mathbf{Y}_{i,j}}
\end{equation}
where $\mathbf{X}$ is the confidence map, and the $\mathbf{Y}$  is the ground truth image.

\section{Experiment}\label{Section:Experiment}
\begin{table*}
	\renewcommand\arraystretch{1.2}
	\footnotesize
	\centering
	\caption{Comparison with Other State-of-the-art methods\dag $\;$on three datasets. The best and second results are in \textcolor{red}{\textbf{red}} and \textcolor{blue}{\textbf{blue}}, respectively. The metrics considered include IoU ($10^{-2}$), $\text{nIoU}$ ($10^{-2}$), $P_d$ ($10^{-2}$), $F_a$ ($10^{-6}$). }
	\label{tab:sota}
	\setlength{\tabcolsep}{3.pt}
	\begin{tabular}{l|c|c|c|cccc|cccc|cccc}
		\noalign{\hrule height 1pt}
		\multirow{2}{*}{Method} & \multirow{2}{*}{Publish} & \multirow{2}{*}{Param\ddag} & \multirow{2}{*}{FLOPs} & \multicolumn{4}{c|}{IRSTD-1k}   & \multicolumn{4}{c|}{SIRSTAUG}   & \multicolumn{4}{c}{NUDT-SIRST} \\ \multicolumn{1}{l|}{} & \multicolumn{1}{l|}{}  &\multicolumn{1}{l|}{} & \multicolumn{1}{l|}{} 
		& IoU $\uparrow$ &  nIoU $\uparrow$ & $P_d$ $\uparrow$ & $F_a$ $\downarrow$ & IoU $\uparrow$ &  nIoU $\uparrow$ & $P_d$ $\uparrow$ & $F_a$ $\downarrow$ & IoU $\uparrow$ &  nIoU $\uparrow$ & $P_d$ $\uparrow$ & $F_a$ $\downarrow$ \\
		\noalign{\hrule height 1pt}
		MDvsFA \cite{wang2019miss}                  & ICCV'19  & 3.919M                         & 998.6G  & 49.50                     & 47.41  & 82.11                     & 80.33 &  $\textemdash$ &  $\textemdash$ &  $\textemdash$ &  $\textemdash$ & 75.14 &  $\textemdash$ & 90.47 & 25.34 \\
		ACM  \cite{Dai_2021_WACV}                     & WACV'21 & 0.5198M                        & 2.009G  & 63.39                     & 60.81  & 91.25                     & 8.961 & 73.84  & 69.83  & 97.52  & 76.35  & 68.48                     & 69.26  & 96.26  & 10.27  \\
		ALCNet  \cite{ALCNet}                  & TGRS'21 & 0.5404M                        & 2.127G  & 62.05                     & 59.58  & 92.19                     & 31.56 &  $\textemdash$ &  $\textemdash$ &  $\textemdash$ &  $\textemdash$ & 61.13   & 60.61  & 96.51 & 29.09 \\
		DNANet \cite{DNANet}                   & TIP'22  & 4.6968M                        & 56.08G  & 68.87                     & 67.53  & 94.95                     & 13.38 &  74.88  & 70.23  & 97.80   & 30.07  & 92.67                     & 92.09  & 99.53  & 2.347  \\
		ISNet \cite{ISNet}                    & CVPR'22 & 1.09M                          & 121.90G & 68.77                     & 64.84  & 95.56                     & 15.39 & 72.50   & 70.84  & 98.41  & 28.61  & 89.81                     & 88.92  & 99.12  & 4.211  \\
		AGPCNet \cite{AGPCNet}                  & TAES'23 & 12.36M                         & 327.54G & 68.81                     & 66.18  & 94.26                     & 15.85 & 74.71  & 71.49  & 97.67  & 34.84  & 88.71                     & 87.48  & 97.57  & 7.541  \\
		UIUNet  \cite{UIUNet}                  & TIP'23  & 50.54M                         & 217.85G & 69.13                     & 67.19  & 94.27                     & 16.47 & 74.24  & 70.57  & 98.35  & 23.13  & 90.77                     & 90.17  & 99.29  & 2.39   \\
		RDIAN  \cite{RDIAN}                   & TGRS'23 & 0.131M                         & 14.76G  & 64.37                     & 64.90   & 92.26                     & 18.2  & 74.19  & 69.8   & 99.17  & 23.97  & 81.06                     & 81.72  & 98.36  & 14.09  \\
		MTUNet \cite{MTUNet}                   & TGRS'23 & 4.07M                          & 24.43G  & 67.50                      & 66.15  & 93.27                     & 14.75 & 74.70   & 70.66  & 98.49  & 39.73  & 87.49                     & 87.70   & 98.60   & 3.76   \\
		RepISD-Net \cite{wu2023repisd}                  & TGRS'23 & 0.28M                          & 25.76G  & 65.45                     & $\textemdash$  & 91.59                     & 7.62 & $\textemdash$   & $\textemdash$  & $\textemdash$  & $\textemdash$  & 89.44                     & $\textemdash$   & 98.65   & 6.18   \\
		SRNet  \cite{SRNet}                   & TMM'23  & 0.4045M & $\textemdash$  & 69.45                     & 65.51  & 96.77                    & 13.05 & $\textemdash$ & $\textemdash$ & $\textemdash$ & $\textemdash$ & $\textemdash$                    & $\textemdash$ & $\textemdash$ & $\textemdash$ \\
		IRPruneDet \cite{zhang2024irprunedet}                   & AAAI'24  & 0.1802M                         & 0.938G  & 64.54                    & 62.71  & 91.74                     & 16.04 &  $\textemdash$ &  $\textemdash$ &  $\textemdash$ &  $\textemdash$ & $\textemdash$ &  $\textemdash$ & $\textemdash$ & $\textemdash$ \\
		TCI-Former \cite{chen2024tci}                  & AAAI'24  & 3.66M                         & 5.83G  & 70.14                    & 67.69  & 96.31                     & 14.81 &  $\textemdash$ &  $\textemdash$ &  $\textemdash$ &  $\textemdash$ & $\textemdash$ &  $\textemdash$ & $\textemdash$ & $\textemdash$ \\
		MSHNet  \cite{MSHNet}                  & CVPR'24 & 4.07M                          & 24.43G  & 67.87 & 61.70   & 92.86 & 8.88  & $\textemdash$ & $\textemdash$ & $\textemdash$ & $\textemdash$ & 80.55 & $\textemdash$ & 97.99  & 11.77  \\
		HintHCFNet  \cite{10764792}                  & TGRS'24 & 14.39M  & 23.23G  & 62.24 & 58.87   & 90.17 & 17.12  & $\textemdash$ & $\textemdash$ & $\textemdash$ & $\textemdash$ & 93.29 & 93.67 & 98.93  & \first{0.67}  \\
		$\text{Mrf}^3\text{Net}$ \cite{MRF3Net} & TGRS'24 & 0.538M                         & 33.2G   & 67.79                     & 68.74  & 95.28                     & 14.5  & $\textemdash$ & $\textemdash$ & $\textemdash$ & $\textemdash$ & 95.21                     & 95.23  & 99.36  & 1.86   \\
		SCTransNet \cite{SCTransNet}               & TGRS'24 & 11.19M                         & 67.4G   & 68.03                     & 68.15  & 93.27                     & 10.74 & $\textemdash$ & $\textemdash$ & $\textemdash$ & $\textemdash$ & 94.09                     & 94.38  & 98.62  & 4.29   \\
		MLP-Net \cite{10793117}                  & TGRS'24  & 8.03M                         & 8.714G  & 56.67                    & 57.16  & 92.01                     & 13.12 &  $\textemdash$ &  $\textemdash$ &  $\textemdash$ &  $\textemdash$ & 90.44 &  92.28 & \first{99.73} & 1.16 \\
		MDAFNet  \cite{li2024moderately}                 & TGRS'24  & 10.07M                         & 100.46G  & $\textemdash$                    & 64.25  & 76.66                     & 6.11 &  $\textemdash$ &  $\textemdash$ &  $\textemdash$ &  $\textemdash$ & $\textemdash$ &  93.42 & 95.58 & 1.89 \\
		DCFR-Net \cite{fan2024diffusion}                 & TGRS'24  & $\textemdash$                         & $\textemdash$ & 65.41                    & 65.45  & 93.60                     & 7.34 &  $\textemdash$ &  $\textemdash$ &  $\textemdash$ &  $\textemdash$ & 86.93 &  86.92 & 98.26 & 2.48 \\
		CSRNet \cite{CSRNet}       & TIP'24  & 0.4045M                        & 121G    & 65.87                     & 66.70   & \first{98.16}                     & 12.08 & $\textemdash$ & $\textemdash$ & $\textemdash$ & $\textemdash$ & $\textemdash$                    & $\textemdash$ & $\textemdash$ & $\textemdash$ \\
		OIPF-SCT \cite{ma2025oipf}                  & TAES'25  & 22.37M                         & 80.97G  & 66.60                     & 66.31  & 94.95                     & 12.1 &  $\textemdash$ &  $\textemdash$ &  $\textemdash$ &  $\textemdash$ & \second{95.43} &  \second{95.53} & 99.26 & \second{0.90} \\
		CFD-Net \cite{10949663}                  & TNNLS'25  & 1.05M                         & 110.37G  & 69.97                    & $\textemdash$  & 95.96                     & 14.21 &  $\textemdash$ &  $\textemdash$ &  $\textemdash$ &  $\textemdash$ & $\textemdash$ &  $\textemdash$ & $\textemdash$ & $\textemdash$ \\
		L2SKNet \cite{10813615}                  & TGRS'25  & 0.899M                         & 6.89G  & 67.81                    & $\textemdash$  & 90.24                     & 17.46 &  74.00 &  $\textemdash$ &  99.17 &  54.90 & 93.58 &  $\textemdash$ & 97.57 & 5.33 \\
		IRMamba \cite{zhang2025irmamba}                  & AAAI'25  & 10.51M                         & $\textemdash$  & 70.04                    & $\textemdash$  & 95.81                     & 5.92 &  $\textemdash$ &  $\textemdash$ &  $\textemdash$ &  $\textemdash$ & 95.18 &  $\textemdash$ & 99.26 & 1.309 \\
		PConv \cite{yang2025pinwheel}                  & AAAI'25  & 2.54M                         & 18.6G  & 67.45                   & $\textemdash$  & 92.20                     & 10.70 &  $\textemdash$ &  $\textemdash$ &  $\textemdash$ &  $\textemdash$ & $\textemdash$ &  $\textemdash$ & $\textemdash$ & $\textemdash$ \\
		MMLNet \cite{li2025multi}                   & TGRS'25  & 3.58M                         & 81.64G  & 67.21                   & $\textemdash$  & 94.28                     & 14.00 &  $\textemdash$ &  $\textemdash$ &  $\textemdash$ &  $\textemdash$ & 81.81 &  $\textemdash$ & 98.43 & 11.77 \\
		HDNet  \cite{11017756}                 & TGRS'25  & 3.84M                         & 5.96G  & 70.26                   & $\textemdash$  & 94.56                     & 4.33 &  $\textemdash$ &  $\textemdash$ &  $\textemdash$ &  $\textemdash$ & 85.17 &  $\textemdash$ & 98.52 & 2.78 \\
		DWTFreqNet \cite{11159515}                  & TGRS'25  & 31.37M                        & 66.87G  & 65.89                   & 66.07  & 92.93                     & 12.87 &  $\textemdash$ &  $\textemdash$ &  $\textemdash$ &  $\textemdash$ & 94.60 &  94.76 & 99.37 & 2.919 \\
		\hline
		\multicolumn{16}{l}{\textit{Deep Unfolding-Based Methods}}  \\
		\hline
		RPCANet  \cite{RPCANet}                 & WACV'24 & 0.68M                          & 179.74G  & 63.21                     & 64.27 & 88.31                     & 43.9  & 72.54  & $\textemdash$ & 98.21  & 34.14  & 89.31                     & 89.03 & 97.14  & 28.7   \\
		DRPCA-Net \cite{11079687}                  & TGRS'25 & 1.169M                          & -  & 64.14                     & - & 92.09                     & 17.92  & \second{76.50}  & $\textemdash$ & 98.62  & 30.60  & 94.16                     & - & 98.41  & 2.55   \\
		\noalign{\hrule height 1pt} \rowcolor[rgb]{0.9,0.9,0.9}
		\multicolumn{16}{l}{\textit{Fine-tuning Models Based on Segment Anything Model} \cite{SAM}}   \\
		\hline
		IRSAM  \cite{IRSAM}                   & ECCV'24 & 12.33M                &286.52G     & 73.69                     & \second{68.97}  & 96.92                     & 7.55  & $\textemdash$ & $\textemdash$ & $\textemdash$ & $\textemdash$ & 92.59                     & 93.29  & 98.87  & 6.94  \\
		
		MDSAM  \cite{gao2024multi}                 & MM'24 & 100.21M          & 123.44G         & 60.82                     & 62.12 & 93.60                     & 19.04  & \first{76.51}  & \second{73.13} & \first{99.86}  & 5.07  & 89.22                     & 92.50 & 99.35  & 10.76   \\
		\rowcolor[rgb]{0.9,0.9,0.9}
		SAM2-UNet  \cite{xiong2024sam2}                 & ICCVW'25 & 4.269M       &203.73G            & 67.10                     & 62.87 & 93.93                     & \second{3.20}  & 74.35  & 71.32 & \first{99.86}  & \first{3.22}  & 70.51                     & 75.18 & 98.30  & 4.31   \\
		SAIST \cite{zhang2025saist}                & CVPR'25 & 389.57M            & $\textemdash$  & 72.14                     & 68.15  & 96.18                     & 4.76  & $\textemdash$  & $\textemdash$ & $\textemdash$  & $\textemdash$  & 95.23                     & 80.42 & 99.28  & 1.31   \\ 
		SAM-SPL \cite{11172325}               & TGRS'25 & 12.25M            & 58.50G  & \second{74.09}                     & 66.06  & 92.59                    & 9.28  & $\textemdash$  & $\textemdash$ & $\textemdash$  & $\textemdash$  & 94.63                     & $\textemdash$ & 99.47  & 2.55   \\ \rowcolor[rgb]{0.9,0.9,0.9}
		\cite{xiong2024sam2} w. Mona \cite{yin20255} & CVPR'25  & 4.668M & 216.48G  & 68.34                     & 63.27 & 94.33                     & 9.88  & 73.95  & 71.19 & 99.79  & 7.99  & 81.11                    & 77.67 & 97.81  & 3.21   \\
		\noalign{\hrule height 1pt} \rowcolor[rgb]{0.9,0.9,0.9}
		\cite{xiong2024sam2} w. Ours & $\textemdash$  & 4.719M &238.17G  &  \textbf{\textcolor{red}{74.78}} & \textbf{\textcolor{red}{69.73}} & \second{98.10} & \textbf{\textcolor{red}{1.3}} & 75.79 & \first{74.23} & \second{99.81} & \second{4.77} & \first{96.76} & \first{96.43} & \second{99.58} & 2.03 \\
		\noalign{\hrule height 1pt}
		\multicolumn{16}{l}{\footnotesize{(\dag) To ensure consistency in the comparisons, we report the results of the Pre-SOTA algorithms as presented in their original papers whenever possible.}}\\
		\multicolumn{16}{l}{\footnotesize{(\ddag) Here, the number of parameters refers to the amount of learnable parameters in the model.}}
	\end{tabular}
\end{table*}
\vspace{-4pt} 
\subsection{Experimental Setup}
\subsubsection{Datasets}  We evaluate the effectiveness of our method across multiple public datasets spanning various downstream tasks. For clarity and conciseness, we present results on the infrared small target detection task in the main text.
For the infrared small target detection task, we conduct evaluations on three widely used benchmark datasets (IRSTD-1k \cite{ISNet}, SIRSTAUG \cite{AGPCNet}, and NUDT-SIRST \cite{DNANet}). These datasets are either characterized by significant cross-scale variations or predominantly contain small targets, and all of them pose challenges due to complex background interference.
\subsubsection{Evaluation Metrics} For the infrared small target detection task, we employ four metrics: IoU, nIoU, $P_d$, and $F_a$. Among these, IoU evaluates the model’s capability in segmenting large targets, while nIoU, a metric specifically designed for infrared small target detection, assesses the segmentation performance on small targets \cite{Dai_2021_WACV}. Together, these metrics comprehensively reflect the model's adaptability to multi-scale target detection.

\subsubsection{Implementation Details} For the infrared small target detection task, both the proposed method and the baseline \cite{xiong2024sam2} use identical settings, based on PyTorch 2.4.0 with CUDA 12.1 acceleration. The network is trained using the Adam optimizer with an initial learning rate of 1e-4, betas (0.9, 0.999), eps 1e-8, weight decay 0 on an NVIDIA A100 GPU. Learning rate decay follows the poly policy \cite{RPCANet}, with a batch size of 8, input size of 512$\times$512 and a maximum of 400 epochs.

\subsection{Comparison with State-of-the-Arts}
\subsubsection{Quantitative Evaluation}\footnote{Details of the cross-dataset evaluation and FPS comparisons can be found in the \underline{\textbf{supplementary material}}.} As summarized in Table \ref{tab:sota}, our SAM fine-tuning approach achieves the best performance in handling cross-scale target variations. Methods that incorporate handcrafted priors during fine-tuning generally outperform task-specific techniques; however, they still struggle with small-scale targets, leading to relatively low nIoU scores. Conventional fine-tuning methods perform poorly across the board, regardless of target size.

IRSAM \cite{IRSAM} leverages handcrafted priors in the wavelet domain based on the assumption of significant temperature contrasts at target boundaries. While effective for targets with pronounced thermal differences, its performance deteriorates in complex backgrounds where boundary cues are obscured. MDSAM \cite{gao2024multi} enhances fine-tuning with multi-scale and detail modules but lacks task-specific design, limiting its ability to mitigate the model’s intrinsic texture bias and resulting in suboptimal detection of weakly textured targets. Similarly, SAM2-UNet \cite{xiong2024sam2} struggles under cluttered backgrounds, showing notable performance drops on the NUDT-SIRST dataset \cite{DNANet}, and falls short even compared to specialized methods. SAIST \cite{zhang2025saist} incorporates multimodal inputs with scene textual descriptions, which are often impractical to obtain, and suffers from increased model complexity due to CLIP integration.
\begin{figure}
	\centering
	\includegraphics[width=\linewidth]{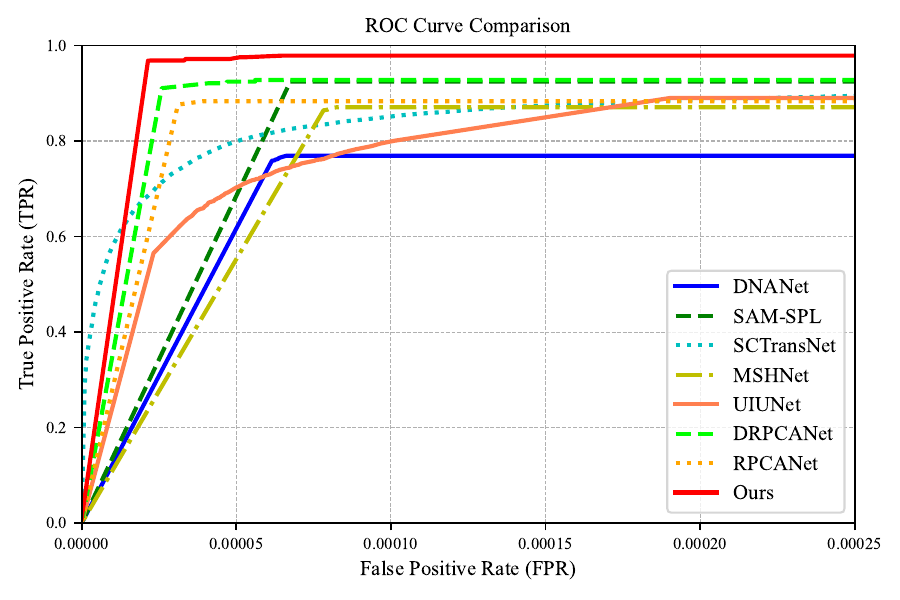}
	\caption{ROC curves of ours and other approaches on IRSTD-1k \cite{ISNet}.}
	\label{fig:ROC}   
\end{figure}
Distinct from these approaches, our method avoids task-specific handcrafted features altogether. Instead, by introducing a shape bias representation, we effectively counteract the model’s texture bias, thereby achieving superior performance in infrared small target detection. As shown in Fig. \ref{fig:ROC}, we further present the ROC curve on the IRSTD-1K dataset \cite{ISNet} to validate the effectiveness of the proposed method.
\subsubsection{Qualitative Evaluation} 
\begin{figure*}
	\centering
	\includegraphics[width=\linewidth]{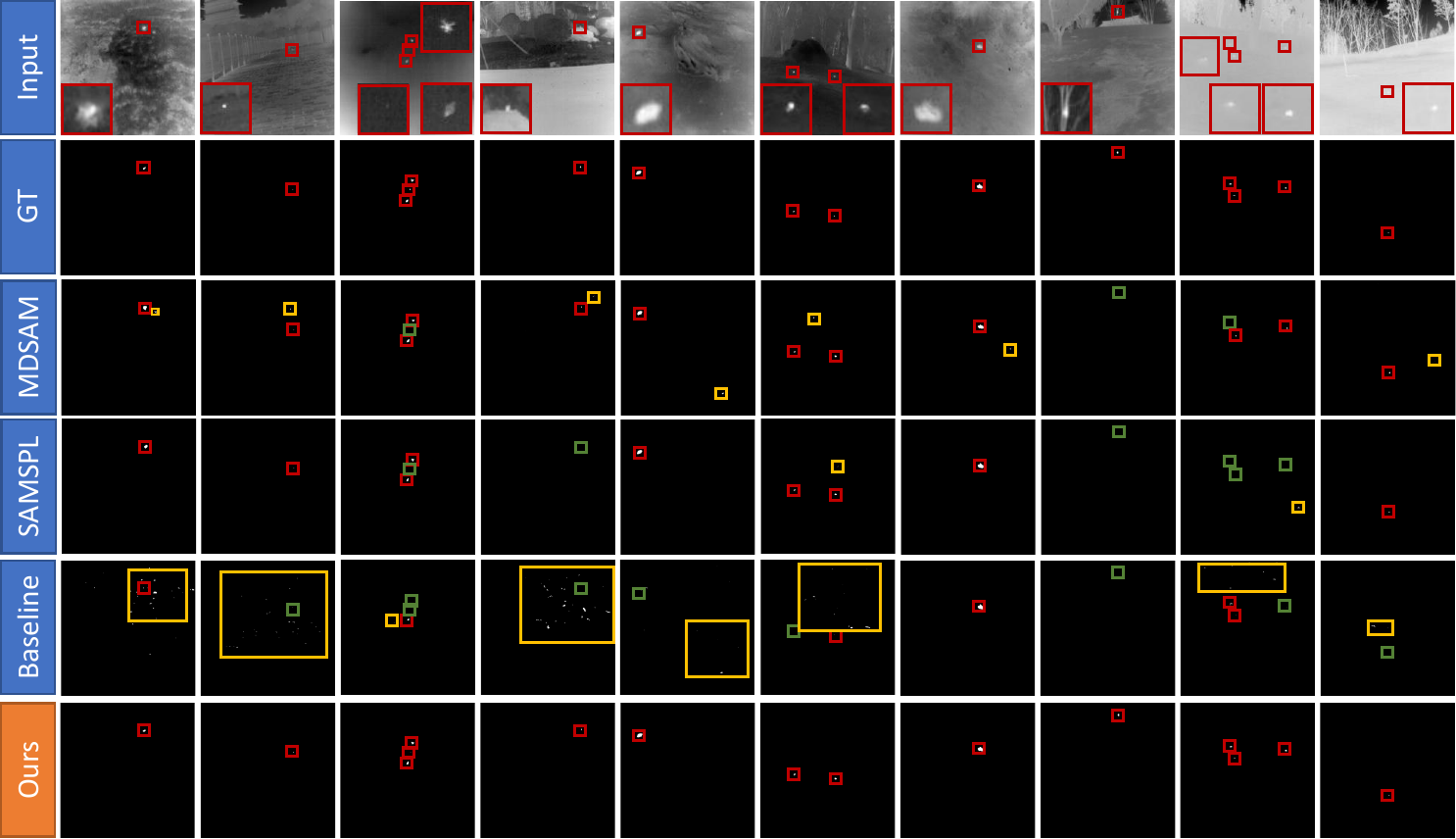}
	\caption{Visualization comparison of detection results between Ours and the other fine-tuning models based on SAM across various challenging scenarios. Correct detections, false alarms, and missed detections are indicated by \textcolor{red}{red}, \textcolor[rgb]{1.0, 0.84, 0.0}{yellow}, and \textcolor{green}{green} bounding boxes, respectively.}
	\label{fig:visual_more}   
\end{figure*}
As shown in Fig. \ref{fig:visual_more}, the proposed method demonstrates superior performance across various challenging scenarios. It effectively adapts to targets of varying sizes and signal-to-clutter ratios. Notably, compared to our baseline, the method exhibits enhanced robustness under texture interference, which validates the effectiveness of introducing shape bias into the model.

\subsection{Ablation Study}
To validate the effectiveness of the proposed approach, this section presents ablation studies analyzing the contribution of each component to the overall performance. Our baseline is SAM2-UNet \cite{xiong2024sam2}. For a fair comparison, all hyperparameter settings are kept consistent with the baseline.
\begin{table}[]
	\centering
	\caption{Ablation study on the effectiveness of each component. The best and second results are in \textcolor{red}{\textbf{red}} and \textcolor{blue}{\textbf{blue}}, respectively.}
	\label{Tab:LM}
	\begin{tabular}{cc|cccc}
		\toprule
		\multicolumn{2}{c|}{Variants} & \multicolumn{4}{c}{NUDT-SIRST} \\
		\multicolumn{1}{c|}{L\dag}                        & S\ddag  & IoU ($10^{-2}$)   & nIoU ($10^{-2}$)  & $P_d$ ($10^{-2}$)    & $F_a$ ($10^{-6}$)   \\ \midrule
		\multicolumn{1}{c|}{\XSolidBrush}   & \XSolidBrush  & 70.51  & 75.18  & \second{98.30}  & \second{4.31} \\
		\multicolumn{1}{c|}{\CheckmarkBold}   & \XSolidBrush  & 43.35  & 40.71  & 66.57  & 133.1 \\
		\multicolumn{1}{c|}{\XSolidBrush}   & \CheckmarkBold  & \second{85.31}  & \second{83.37}  & 90.84  & 77.6 \\
		\multicolumn{1}{c|}{\CheckmarkBold}   & \CheckmarkBold  & \first{96.76}  & \first{96.43}  & \first{99.58} & \first{2.03} \\ \bottomrule
		\multicolumn{6}{l}{\footnotesize{(\dag) L represents the normalization layers subject to fine-tuning.}}\\
		\multicolumn{6}{l}{\footnotesize{(\ddag) S denotes the proposed side-tuning sub-network.}}
	\end{tabular}
\end{table} 
\begin{table}[]
	\centering
	\caption{Ablation study on the proposed HDConv Block.}
	\label{Tab:HDConvBlock}
	\begin{tabular}{cc|cccc}
		\toprule
		\multicolumn{2}{c|}{Variants} & \multicolumn{4}{c}{NUDT-SIRST} \\
		\multicolumn{1}{c|}{A\dag}                        & M\ddag  & IoU ($10^{-2}$)   & nIoU ($10^{-2}$)  & $P_d$ ($10^{-2}$)    & $F_a$ ($10^{-6}$)   \\ \midrule
		\multicolumn{1}{c|}{\XSolidBrush}   & \XSolidBrush  & \second{70.51}  & \second{75.18}  & \second{98.30}  & \second{4.31} \\
		\multicolumn{1}{c|}{\CheckmarkBold}   & \XSolidBrush  & 57.69  & 41.58  & 87.40  & 80.8 \\
		\multicolumn{1}{c|}{\XSolidBrush}   & \CheckmarkBold  & 50.43  & 43.97  & 88.61  & 153.9 \\
		\multicolumn{1}{c|}{\CheckmarkBold}   & \CheckmarkBold  & \first{96.76}  & \first{96.43}  & \first{99.58} & \first{2.03} \\ \bottomrule
		\multicolumn{6}{l}{\footnotesize{(\dag) A denotes the proposed Shape Enhanced Large-Kernel Attention.}}\\
		\multicolumn{6}{l}{\footnotesize{(\ddag) M denotes the residual module in the HDConv block.}}
	\end{tabular}
\end{table}
\begin{table}[]
	\centering
	\caption{Ablation study on the effectiveness of ERF'shape.}
	\label{Tab:ERF}
	\begin{tabular}{l|cccc}
		\toprule
		\multirow{2}{*}{Variants} & \multicolumn{4}{c}{NUDT-SIRST} \\
		& IoU   & nIoU ($10^{-2}$)  & $P_d$ ($10^{-2}$)    & $F_a$ ($10^{-6}$)   \\ \midrule
		RepLK \cite{ding2022scaling}                    & \second{89.46} & \second{90.21}  & \second{97.32}  & 6.91 \\
		WTConv \cite{finder2025wavelet}                   & 75.75  & 71.98  & 94.71  & \first{1.82}\\
		SLaKConv \cite{DBLP:conf/iclr/LiuCCCXWKPMW23}                   & 83.33  & 79.42  & 93.65  & 7.99 \\
		Ours                      & \first{96.76}  & \first{96.43}  & \first{99.58} & \second{2.03} \\ \bottomrule
	\end{tabular}
\end{table}
\begin{table}[]
	\centering
	\caption{Ablation study on the proposed SConv.}
	\label{Tab:SConv}
	\begin{tabular}{cccc|cccc}
		\toprule
		\multicolumn{4}{c|}{Variants \dag \ddag}                                                & \multicolumn{4}{c}{NUDT-SIRST} \\ 
		\multicolumn{1}{c|}{S} & \multicolumn{1}{c|}{C} & \multicolumn{1}{c|}{L} & D & IoU   & nIoU  & $P_d$ ($10^{-2}$)    & $F_a$ ($10^{-6}$)   \\ \midrule
		\multicolumn{1}{c|}{\CheckmarkBold}  & \multicolumn{1}{c|}{\XSolidBrush}  & \multicolumn{1}{c|}{\XSolidBrush} & \XSolidBrush & 84.32  & 82.76  & 93.12  & \first{1.73} \\
		\multicolumn{1}{c|}{\CheckmarkBold}  & \multicolumn{1}{c|}{\CheckmarkBold}  & \multicolumn{1}{c|}{\XSolidBrush} & \XSolidBrush & 89.98  & 92.03  & 96.41  & 11.71 \\
		\multicolumn{1}{c|}{\CheckmarkBold}  & \multicolumn{1}{c|}{\CheckmarkBold}  & \multicolumn{1}{c|}{\CheckmarkBold} & \XSolidBrush & 92.56  & 91.17  & 94.95 & \second{1.97} \\
		\multicolumn{1}{c|}{\CheckmarkBold}  & \multicolumn{1}{c|}{\CheckmarkBold}  & \multicolumn{1}{c|}{\XSolidBrush} & \CheckmarkBold & \first{96.76}  & \first{96.43}  & \first{99.58} & 2.03 \\ \bottomrule
		\multicolumn{8}{l}{\footnotesize{(\dag) S denotes small-kernel convolution and C denotes short connections.}} \\
		\multicolumn{8}{l}{\footnotesize{(\ddag) L denotes large-kernel convolution and D denotes decomposition of L.}}
	\end{tabular}
\end{table}
\begin{table}[]
	\centering
	\caption{Ablation study on the proposed SELKA.}
	\label{tab:SELKA}
	\begin{tabular}{ccc|cccc}
		\toprule
		\multicolumn{3}{c|}{Variants \dag  ¶ \ddag}                                                & \multicolumn{4}{c}{NUDT-SIRST} \\ 
		\multicolumn{1}{c|}{S} & \multicolumn{1}{c|}{C} & \multicolumn{1}{c|}{L} & IoU   & nIoU  & $P_d$ ($10^{-2}$)    & $F_a$ ($10^{-6}$)   \\ \midrule
		\multicolumn{1}{c|}{\CheckmarkBold}  & \multicolumn{1}{c|}{\XSolidBrush}  & \multicolumn{1}{c|}{\XSolidBrush}  & 92.11  & 92.16  & 96.68  & 5.99 \\
		\multicolumn{1}{c|}{\CheckmarkBold}  & \multicolumn{1}{c|}{\CheckmarkBold}  & \multicolumn{1}{c|}{\XSolidBrush}  & \second{94.35}  & \second{93.26}  & \second{98.83}  & \first{1.99} \\
		\multicolumn{1}{c|}{\CheckmarkBold}  & \multicolumn{1}{c|}{\CheckmarkBold}  &  \multicolumn{1}{c|}{\CheckmarkBold}  & \first{96.76}  & \first{96.43}  & \first{99.58} & \second{2.03} \\ \bottomrule
		\multicolumn{7}{l}{\footnotesize{(\dag) S denotes the upsampling and downsampling operator.}} \\
		\multicolumn{7}{l}{\footnotesize{(\dag) ¶ represents the module responsible for cross-branch feature interaction.}} \\
		\multicolumn{7}{l}{\footnotesize{(\ddag) denotes the lightweight spatial attention module.}}
	\end{tabular}
\end{table}

\subsubsection{Ablation Study on the Effectiveness of Each Component} 
The contributions of each component are summarized in Table~\ref{Tab:LM}. As the RFB module and its hyperparameters are inherited from the baseline, no ablation is conducted. Fine-tuning only the normalization layers yields limited gains, as it does not modify the model’s perceptual biases. In contrast, tuning only the auxiliary branch brings substantial improvements due to its shape bias. Jointly fine-tuning both achieves the best results, as updated normalization better aligns features with the auxiliary branch.

\subsubsection{Ablation Study on the Proposed HDConv Block} The proposed HDConv Block comprises two parts: a shape-enhanced large-kernel attention for capturing shape bias and a residual block for learning task-specific knowledge. As shown in Table~\ref{Tab:HDConvBlock}, removing either component leads to notable performance drops. Without the residual block, the model lacks adaptability to downstream tasks. Conversely, the attention module alone can highlight discriminative features but fails to leverage them effectively for adaptation.

\subsubsection{Ablation Study on the Gaussian-distributed effective receptive field} \label{Sec:Gau} To validate the effectiveness of the Gaussian-distributed ERF, we compare it against several widely adopted large-kernel convolution variants. The results, presented in Tab.~\ref{Tab:ERF}, align with the ERF patterns illustrated in Fig.~\ref{fig:OverallofERF}. Specifically, when the distribution of the ERF is misaligned with the structural characteristics of the target, target features tend to become entangled with background features. This misalignment introduces two key issues: on one hand, it increases the risk of overfitting during training; on the other hand, the persistent feature entanglement degrades the network’s ability to distinguish small targets in deeper layers. Notably, the smaller the target, the more severe the degradation, resulting in significantly lower nIoU scores.

\subsubsection{Ablation Study on the Proposed SConv} The ablation results of SConv are presented in Table \ref{Tab:SConv}, which are consistent with the theoretical analysis in Section \ref{Sec:The}, thereby validating the effectiveness of both the design and the analysis. Without skip connections, merging small kernels into large ones leads to reduced sensitivity to small objects. Moreover, employing square large-kernel convolutions tends to cause confusion between object features and background features. In addition, it can be observed that incorporating large-kernel convolutions with a large effective receptive field enhances the suppression of false positives. This is attributed to the increased low-pass filtering \cite{11146868} effect of the model as the effective receptive field expands.

\subsubsection{Ablation Study on the Proposed Shape Enhanced Large-kernel Attention} To validate the effectiveness of the proposed attention mechanism, we conduct an ablation study on its internal components. As shown in Tab. \ref{tab:SELKA}, removing the cross-branch feature interaction leads to a performance drop. This is because such interactions help form a more stable Gaussian-shaped effective receptive field. Similarly, removing the spatial attention module also degrades performance, as it eliminates the mechanism’s ability to adaptively select appropriate receptive fields. In its absence, the model falls back to a static weighted averaging via 1$\times$1 convolution, which results in suboptimal performance.

\subsection{Limitations}
\begin{figure}
	\centering
	\includegraphics[width=\linewidth]{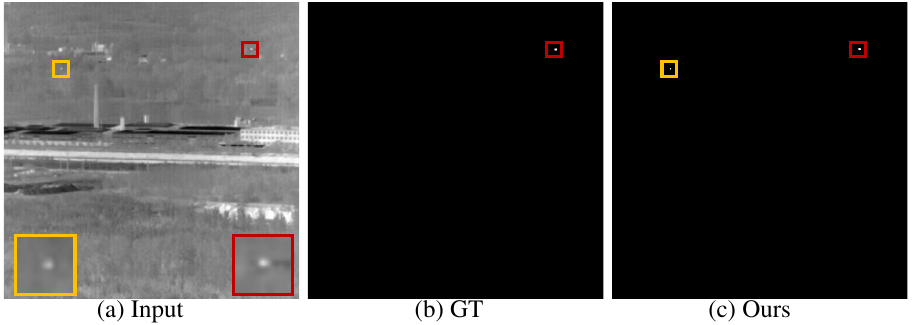}
	\caption{Typical failure cases. The inherent ambiguity of single-frame images makes it challenging to distinguish between dead pixel noise and actual targets.}
	\label{fig:Failed}
\end{figure}
As illustrated in Fig.~\ref{fig:Failed}, the inherent feature ambiguity of single-frame imagery makes it challenging to reliably distinguish between true targets and dead pixel noise. Since dead pixel noise is often correlated with the imaging hardware, its spatial location tends to remain fixed, making it possible to accurately differentiate it using temporal information across frames. To address this limitation, we plan to explore the integration of temporal cues~\cite{zhang2025beyond} in future work.
\section{Conclusion}\label{Section:Conclusion}
In this paper, we propose the Ladder Shape-Biased Side-Tuning method, which introduces shape bias to overcome the inherent texture bias in foundational visual models, thereby enhancing performance on tasks such as infrared small target detection. Unlike previous approaches that equate shape with contour, we argue that shape and contour are not equivalent; shape encompasses both the contour and the internal distribution, thus expanding the concept of shape bias. Finally, we provide a theoretical analysis of the proposed module design to strengthen its rigor and credibility. Our method lays a foundational theoretical basis for improving the performance of foundational visual models on downstream texture-insensitive tasks.

\bibliographystyle{IEEEtran}
\bibliography{reference} 
\end{document}


\title{``LSP-ST: Ladder Shape-Biased Side-Tuning for \\Robust Infrared Small Target Detection'': \\ Supplementary Material}

\author{Guoyi~Zhang~\orcidlink{0009-0004-2931-6698}, Siyang~Chen~\orcidlink{0009-0008-9465-6647}, Guangsheng~Xu~\orcidlink{0009-0006-4235-9398}, Han~Wang~\orcidlink{0009-0008-2794-2520
	},~Donghe~Wang,~and~Xiaohu~Zhang~\orcidlink{0000-0003-4907-1451}
}

\markboth{Journal of \LaTeX\ Class Files,~Vol.~14, No.~8, August~2021}%
{Guoyi Zhang \MakeLowercase{\textit{et al.}}: LSP-ST: Ladder Shape-Biased Side-Tuning for Robust Infrared Small Target Detection}


\maketitle

\begin{abstract}
This supplementary material serves two primary purposes: (1) to report additional experimental results that provide further validation of the findings presented in Section III-C-4 of the main paper; and (2) to analyze the cross-dataset generalization capability of LSP-ST in infrared small target detection tasks, along with a comparative evaluation of its runtime performance in terms of FPS.
\end{abstract}

\section{Experimental Setup}
\subsection{Overview}
Before presenting the results, it is important to note that many task-specific methods often rely on non-generalizable, task-tailored learning strategies (e.g., joint training of shadow detection and removal) for particular downstream tasks, even when fine-tuning vision foundation models. Both our method and the baseline~\cite{xiong2024sam2} aim to explore how to fully unlock the potential of vision foundation models. To this end, we have deliberately avoided such specialized strategies and instead adopted the simplest and most universally applicable approaches across all tasks. Although this choice may result in performance degradation on certain specific tasks, our primary focus remains on the performance differences between the proposed method and the baseline.

Furthermore, we provide an extended discussion on infrared small target detection, including comparisons with state-of-the-art (SOTA) methods in terms of frames per second (FPS) and cross-dataset generalization performance. These additional results further validate the adaptability and efficiency of the proposed method. 

The supplementary material is organized as follows. Section \ref{Sec:Task} introduces task settings and dataset splits, Section \ref{Sec:Details} describes training strategies and resource consumption; Section \ref{Sec:IN} reports results for texture-insensitive tasks, Section \ref{Sec:De} presents results for texture-dependent tasks, Section \ref{Sec:IR} provides extended discussion on infrared small target detection.
Section \ref{Sec:Fu} elaborates on potential future work, drawing upon details provided in the supplementary material.
\subsection{Tasks and Datasets} \label{Sec:Task}
\begin{table}[t]
	\centering
	\caption{Detailed information of datasets for different texture-insensitive tasks.}
	\label{tab:dataset}
	\renewcommand\arraystretch{1.2}
	\renewcommand\tabcolsep{2pt}
	\begin{tabular}{l|c|cc}
		\noalign{\hrule height 1pt}
		Tasks & Dataset  & Train Set & Test Set \\
		\noalign{\hrule height 1pt}
		& CAMO~\cite{le2019anabranch} & 1,000 & 250 \\
		& COD10K~\cite{fan2020camouflaged} & 3,040 & 2,026 \\
		& CHAMELEON~\cite{skurowski2018animal} & - & 76 \\
		\multirow{-4}{*}{Camouflaged Object Detection} & NC4K~\cite{lv2021simultaneously} & - &  4,121 \\
		\hline
		& MAS3K~\cite{li2021marine} & 1,769 & 1,141 \\
		\multirow{-2}{*}{Marine Animal Segmentation} & RMAS~\cite{fu2023masnet} & 2,514 & 500 \\
		\hline
		& MSD~\cite{yang2019my} & 3,063 & 955 \\
		\multirow{-2}{*}{Mirror Detection} & PMD~\cite{lin2020progressive} & 5,096 & 571 \\
		\hline
		\multirow{-1}{*}{Shadow Detection} & ISTD~\cite{wang2018stacked} & 5,096 & 571 \\
		\noalign{\hrule height 1pt}
	\end{tabular}
\end{table}
\begin{table}[t]
	\centering
	\caption{Detailed information of datasets for texture-dependent tasks.}
	\label{tab:dataset1}
	\renewcommand\arraystretch{1.2}
	\renewcommand\tabcolsep{2pt}
	\begin{tabular}{l|c|cc}
		\noalign{\hrule height 1pt}
		Tasks & Dataset  & Train Set & Test Set \\
		\noalign{\hrule height 1pt}
		& DUTS~\cite{wang2017learning} & 10,553 & 5,019 \\
		& DUT-OMRON~\cite{yang2013saliency} & - & 5,168\\
		& HKU-IS~\cite{li2015visual} & - & 4,447\\
		& PASCAL-S~\cite{li2014secrets} & - & 850 \\
		\multirow{-5}{*}{Salient Object Detection} & 
		ECSSD~\cite{yan2013hierarchical} & - & 1,000\\
		\noalign{\hrule height 1pt}
	\end{tabular}
\end{table}
We categorize the tasks into two major types: texture-insensitive tasks, in which texture serves as a source of interference for object segmentation; and texture-dependent tasks, in which texture acts as a helpful cue for object segmentation. 
The specific tasks and dataset settings for texture-insensitive tasks are summarized in Tab. \ref{tab:dataset}, while texture-dependent tasks include only salient object detection \cite{9320524}, with details provided in Tab. \ref{tab:dataset1}.

\subsection{Evaluation Metrics} For the evaluation of each task, we employ the standard metrics most widely adopted in the literature, as following:
\begin{itemize}
	\item Camouflaged Object Detection: S-measure ($S_\alpha$), adaptive F-measure ($F_{\beta}$), mean E-measure ($E_{\phi}$), and mean absolute error (MAE).
	\item Salient Object Detection: S-measure ($S_\alpha$), mean E-measure ($E_{\phi}$), and mean absolute error (MAE).
	\item Marine Animal Segmentation: mIoU, S-measure ($S_\alpha$), weighted F-measure ($F_{\beta}^{w}$), mean E-measure ($E_{\phi}$), and mean absolute error (MAE).
	\item Mirror Detection: IoU, Acc, F-measure, and mean absolute error (MAE), and Balance Error Rate (BER). 
	\item Shadow Detection: Balance Error Rate (BER).
\end{itemize}

\subsection{Implementation Details} \label{Sec:Details} Consistent with the baseline \cite{xiong2024sam2}, our method is implemented using PyTorch 2.4 and trained on an NVIDIA A100 GPU, with 40GB memory. We adopt the AdamW optimizer with an initial learning rate of 0.001, and apply cosine decay to stabilize training. Two data augmentation strategies are used: random vertical and horizontal flipping. Unless otherwise specified, we use the Hiera-L variant of SAM2. All input images are resized to 352$\times$352, and the batch size is set to 12. The number of training epochs is set to 50 for camouflaged object detection and salient object detection, and to 20 for marine animal segmentation and mirror detection.
The loss function is also consistent with the baseline, where $\mathcal{L} = \mathcal {L}_{IoU}^w + \mathcal {L}_{BCE}^w$.
Additionally, deep supervision is applied to all segmentation outputs $S_i$.

The comparison of resource consumption for baseline, Mona fine-tuning, and our proposed method with an input size of 352$\times$352 and batch size of 1 is summarized in Tab. \ref{tab:time}. Although our method demonstrates superior performance, it incurs slightly higher resource usage.

\begin{table}[]
	\centering
	\caption{Comparison of resource consumption among different methods}
	\label{tab:time}
	\begin{tabular}{@{}c|c|c@{}}
		\toprule
		Methods                        & Inference time        & Inference GPU memory usage\dag \\ \midrule
		\multicolumn{1}{l|}{SAM2-UNet \cite{xiong2024sam2}} & \multicolumn{1}{c|}{39.669ms} &   1680MB                         \\
		\multicolumn{1}{l|}{\cite{xiong2024sam2} w. Mona \cite{yin20255}}      & \multicolumn{1}{c|}{46.618ms} &   1782MB                         \\
		\multicolumn{1}{l|}{Ours}      & \multicolumn{1}{c|}{51.708ms} &   2166MB                         \\ \bottomrule
		\multicolumn{3}{l}{\footnotesize{(\dag) More memory-efficient than Bidirectional lateral connections \cite{sung2022lst,diao2024unipt}.}}\\
	\end{tabular}
\end{table}

\subsection{Texture-insensitive Tasks} \label{Sec:IN}
\begin{table*}[h!]
	\renewcommand\arraystretch{1.2}
	\footnotesize
	\centering
	\caption{Comparison on marine animal segmentation and shadow detection. The best and second results are in \textcolor{red}{\textbf{red}} and \textcolor{blue}{\textbf{blue}}, respectively.}
	\vspace{-10pt}
	\label{tab:Marine_Animal_Segmentation}
	\subfloat{
		\begin{minipage}[t]{0.78\textwidth}
			\setlength{\tabcolsep}{3.5pt}
			\scalebox{1.0}{\begin{tabular}{l|c|ccccc|ccccc}
					\noalign{\hrule height 1pt}
					\multirow{2}{*}{Method}  & \multirow{2}{*}{Publish} &  \multicolumn{5}{c|}{MAS3K} & \multicolumn{5}{c}{RMAS} \\ \multicolumn{1}{l|}{} & \multicolumn{1}{l|}{}
					&mIoU $\uparrow$ & $S_\alpha$ $\uparrow$ & $F_\beta^w $ $\uparrow$&$ E_\phi$ $\uparrow$ & MAE $\downarrow$&mIoU $\uparrow$ & $S_\alpha$ $\uparrow$ & $F_\beta^w $ $\uparrow$& $ E_\phi$ $\uparrow$ & MAE $\downarrow$ \\
					\noalign{\hrule height 1pt}
					ECDNet~\cite{li2021marine}  &TCSVT'21 &0.711&0.850&0.766&0.901&0.036&0.664&0.823&0.689&0.854&0.036\\
					OCENet~\cite{liu2022modeling} &WACV'22 &0.667&0.824&0.703&0.868&0.052&0.680&0.836&0.752&0.900&0.030\\
					ZoomNet~\cite{pang2022zoom}  &CVPR'22 &0.736&0.862&0.780&0.898&0.032&0.728&0.855&0.795&0.915&\textcolor{blue}{\textbf{0.022}}\\
					MASNet~\cite{fu2023masnet} &J-OE'23 &0.742&0.864&0.788&0.906&0.032&0.731&0.862&0.801&0.920&0.024\\
					SAM~\cite{SAM} &ICCV'23 &0.566&0.763&0.656&0.807&0.059&0.445&0.697&0.534&0.790&0.053\\
					SAM-Ad\cite{chen2023sam}  &ICCVW'23 &0.714&0.847&0.782&0.914&0.033&0.656&0.816&0.752&0.927&0.027\\
					SAM-DA~\cite{lai2023detect}  &ARXIV'23 &0.742&0.866&0.806&0.925&0.028&0.686&0.833&0.780&0.926&0.024\\
					MAS-SAM \cite{yan2024mas}  &IJCAI'24 &0.788&0.887&0.840&0.938&0.025&\second{0.742}&0.865&\second{0.819}&\second{0.948}&\first{0.021}\\
					Dual-SAM\cite{zhang2024fantastic} &CVPR'24 &0.789&0.884&0.838&0.933&0.023&0.735&0.860&0.812&0.944&\textcolor{blue}{\textbf{0.022}}\\\rowcolor[rgb]{0.9,0.9,0.9}
					SAM2-UNet \cite{xiong2024sam2} &ICCVW'25 &\textcolor{blue}{\textbf{0.799}}&\textcolor{blue}{\textbf{0.903}}&\textcolor{blue}{\textbf{0.848}}&\textcolor{blue}{\textbf{0.943}}&\textcolor{blue}{\textbf{0.021}}&0.738&\textcolor{blue}{\textbf{0.874}}&0.810&0.944&\textcolor{blue}{\textbf{0.022}}\\\rowcolor[rgb]{0.9,0.9,0.9}
					\cite{xiong2024sam2} w. MONA \cite{yin20255} &CVPR'25 &0.732&0.867&0.788&0.912&0.033&0.658&0.832&0.730&0.908&0.032\\
					\hline
					
					\noalign{\hrule height 1pt} \rowcolor[rgb]{0.9,0.9,0.9}
					Ours &- &\textcolor{red}{\textbf{0.822}}&\textcolor{red}{\textbf{0.911}}&\textcolor{red}{\textbf{0.869}}&\textcolor{red}{\textbf{0.950}}&\textcolor{red}{\textbf{0.020}}&\textcolor{red}{\textbf{0.760}}&\textcolor{red}{\textbf{0.880}}&\textcolor{red}{\textbf{0.829}}&\textcolor{red}{\textbf{0.953}}&\textcolor{red}{\textbf{0.021}} \\
					\noalign{\hrule height 1pt}
			\end{tabular}}
	\end{minipage}}
	\subfloat{
		\begin{minipage}[t]{0.19\textwidth}
			\setlength{\tabcolsep}{3.5pt}
			\scalebox{1.07}{\begin{tabular}{l|c}
					\noalign{\hrule height 1pt}
					\multirow{2}{*}{Method} & \multirow{2}{*}{ISTD} \\[4pt] 
					& $\text{BER}\downarrow$ \\
					\noalign{\hrule height 1pt}
					$\text{EVPv1}_{23}$ \cite{liu2023explicit_CVPR} & 1.35 \\
					$\text{EVPv2}_{23}$ \cite{liu2023explicit} & 1.35 \\
					$\text{SDTR}_{23}$ \cite{wu2023many} & 1.51 \\
					$\text{ShadowSAM}_{23}$ \cite{10315174} & 1.36 \\
					$\text{RMLANet}_{23}$ \cite{10144800} & \second{1.01} \\
					$\text{SADT}_{24}$ \cite{wang2024shadow} & 1.05 \\
					$\text{TS-SAM}_{24}$ \cite{yu2024ts}  & 1.04 \\\rowcolor[rgb]{0.9,0.9,0.9}
					$\text{SAM2-UNet}_{24}$ \cite{xiong2024sam2} & 1.62 \\\rowcolor[rgb]{0.9,0.9,0.9}
					$\text{MONA}_{25}$ \cite{yin20255}  & 2.67 \\
					\noalign{\hrule height 1pt} \rowcolor[rgb]{0.9,0.9,0.9}
					Ours & \first{0.93} \\
					\noalign{\hrule height 1pt}
			\end{tabular}}
	\end{minipage}}
\end{table*}
\begin{table*} [h!]
	\renewcommand\arraystretch{1.2}
	\footnotesize
	\centering
	\caption{Comparison with SOTA methods on  camouflaged object detection. The best and second results are in \textcolor{red}{\textbf{red}} and \textcolor{blue}{\textbf{blue}}, respectively.}
	\label{tab:sota_Camouflaged_Object_Detection}
	\setlength{\tabcolsep}{3.5pt}
	\scalebox{1.0}{\begin{tabular}{l|c|cccc|cccc|cccc|cccc}
			\noalign{\hrule height 1pt}
			\multirow{2}{*}{Method} & \multirow{2}{*}{Publish} & \multicolumn{4}{c|}{CHAMELEON (N=76)}   & \multicolumn{4}{c|}{CAMO (N=250)}   & \multicolumn{4}{c|}{COD10K (N=2026)} & \multicolumn{4}{c}{NC4K (N=4121)} \\ \multicolumn{1}{l|}{} & \multicolumn{1}{l|}{}  
			& $S_\alpha \uparrow$ & $E_\phi\uparrow$ 
			& $F_\beta^\omega\uparrow$ & \multicolumn{1}{c|}{$M\downarrow$}& $S_\alpha \uparrow$ & $E_\phi\uparrow$ 
			& $F_\beta^\omega\uparrow$ & \multicolumn{1}{c|}{$M\downarrow$} & $S_\alpha \uparrow$ & $E_\phi\uparrow$ 
			& $F_\beta^\omega\uparrow$ & \multicolumn{1}{c|}{$M\downarrow$} & $S_\alpha \uparrow$ & $E_\phi\uparrow$ 
			& $F_\beta^\omega\uparrow$ & \multicolumn{1}{c}{$M\downarrow$} \\
			\noalign{\hrule height 1pt}
			\multicolumn{18}{l}{\textit{Task-specific Proprietary Methods}}  \\
			\hline
			
			SARNe\cite{sarnet} &TCSVT'23
			& \textcolor{red}{\textbf{.933}} & \textcolor{red}{\textbf{.978}}
			& \textcolor{red}{\textbf{.909}} & \textcolor{blue}{\textbf{.017}}
			& .874 & .935
			& .844 & .046
			& .885 & .947 
			&.820 & .021
			& .889 & .940
			& .851 & .032
			\\
			
			FSPNe\cite{fspnet} &CVPR'23
			& .908 & .965
			& .851 & .023
			&.856 & .928
			& .799 & .050
			& .851 & .930
			& .735 & .026
			& .878 & .937
			& .816 &.035
			\\
			
			MSCAFNe\cite{mscafnet} &TCSVT'23
			& .912 & .970
			& .865 & .022
			& .873 & .937
			& .828 & .046
			& .865 & .936 
			& .776 &.024
			&.887 & .942
			& .838 & .032
			\\
			
			HitNet\cite{hitnet} &AAAI'23
			& .921 & \textcolor{blue}{\textbf{.972}}
			& .897 & .019
			& .849 & .910 
			& .809 & .055
			& .871 & .938
			& .806 & .023 
			& .875 & .929 
			& .834 & .037
			\\
			ZSCOD\cite{10234216} &TIP'23
			& - & -
			& - & -
			& .823 & .884 
			& .759 & .069
			& .821 & .893
			& .691 & .036
			& - & - 
			& - & -
			\\
			PUENet \cite{10159663} &TIP'23
			& .910 & .957
			& - & .022
			& .877 & .930
			& - & .045
			& .873 & .938
			& - & .022
			& .898 & .945 
			& - & .028
			\\
			FSNet\cite{10103836} &TIP'23
			& .905 & .938
			& - & .022
			& .880 & .933 
			& - & .041
			& .870 & .938
			& - & .023
			& .891 & .940 
			& - & .031
			\\

			Yang \etal \cite{10231131} &TCSVT'24
			& .906 & .966
			& .888 & .024
			& .835 & .902
			& .828 & .063
			& .834 & .912
			& .778 & .030
			& .859 & .922
			& .843 &.041
			\\
			
			Liu \etal \cite{10536093} &TCSVT'24
			& - & -
			& - & - 
			& .886 & .939 
			& .854 & .040
			& .881 & .933
			& .803 & .022 
			&.899 & \textcolor{blue}{\textbf{.945}}
			& .854 & .028
			\\
			
			Wang \etal \cite{10616011} &TCSVT'24
			&.918 & .959
			& .880 & .019
			& .868 & .925 
			& .832 &.047
			& .869 & .936
			&.792 & .023 
			& .887 &.939
			&.846 & .031
			\\
			
			MVGNet \cite{10568109} &TCSVT'24
			& .922 & .969
			& .882 &.019
			& .879 & .930
			& .839 & .045
			& .877 &.936
			& .799 & .022 
			& .894 & .938
			& .850 & .030
			\\
			
			CamoFormer \cite{10623294} &TPAMI'24
			& - & -
			& - &-
			& .878 & .934
			& .839 & .044 
			&.872 & .934
			& .793 & .022 
			& .893 & .940 
			& .850 & .030
			\\
			
			RISNet \cite{wang2024depth} &CVPR'24
			& - & -
			& - & -  
			& .870 & .922
			& .827 & .050 
			& .873 &.931
			&.799 & .025
			& .882 & .925 
			& .834 &.037
			\\
			
			HGINet \cite{10719619} &TIP'24
			& - & -
			& - &-  
			& .874 & .937
			& .848 & .041
			& .882 & \textcolor{red}{\textbf{.949}}
			& .821 &.019
			& .894 &\textcolor{red}{\textbf{.947}}
			& \textcolor{blue}{\textbf{.865}} & \textcolor{blue}{\textbf{.027}}
			\\
			
			IdeNet \cite{10661228} &TIP'24
			& .926 & .967
			& .899 & .018  
			& .848 & .904
			& .802 & .057 
			& .885 &.940
			&.818 & .020
			& .889 & .935 
			& .848 &.032
			\\
			
			SENet \cite{10844040} &TIP'25
			& .918 & .957
			& .878 &.019  
			& \textcolor{blue}{\textbf{.888}} & .932 
			& .847 & .039
			& .865 & .925 
			& .780 & .024 
			& .889 & .933 
			& .843 &.032			
			\\
			
			HUNTNet \cite{11169412} &TIP'25
			& - & -
			& - &-  
			&.883 & .938 
			& - & .042
			& .878 & .932 
			& - & .021 
			& .895 & .945
			& - &.029			
			\\
			
			CFRN \cite{11153753} &TIP'25
			& .910 & .963
			& .871 & .020  
			&.881 & \second{.941} 
			& .849 & .039
			& .872 & .942 
			& .800 & .021 
			& .891 & .945
			& .855 &.029			
			\\
			
			PRBE-Net \cite{10814101} &TMM'25
			& .918 &.951
			& .878 & .020 
			& .876 &.928 
			&.837&.045 
			& .867 & .932
			& .793 &.021 
			&.887 & .931
			& .845 & .031
			
			\\
			
			UTNet \cite{11175541} &TMM'25
			& .918 & .964 
			& .886 & .019  
			& .868 & .915
			& .853 & .046 
			& .868 & .931 
			& .805 & .022 
			& .888 & .932 
			& \second{.865} & .032
			
			\\
			
			EASE \cite{du2025shift} &CVPR'25
			& .864 & .916 
			& .827 & .037  
			& .807 & .865
			& .771 & .078 
			& .866 & .918 
			& .811 & .023 
			& .866 & .915 
			& .833 & .039
			
			\\
			
			AdaptCOD \cite{yin2025camouflaged} &IJCV'25
			& - & -
			& - & -  
			& .886 & .932
			& - & .043 
			& \second{.892} & .938
			& - & .021
			& \textcolor{blue}{\textbf{.906}} & .942
			& - & .029
			
			\\
			
			\noalign{\hrule height 1pt} 
			\multicolumn{18}{l}{\textit{Diffusion Model for  Camouflaged Object Detection}}
			\\ \hline
			
			PrObeD \cite{asnani2023probed} &NIPS'23
			&- & - 
			& - & - 
			& .797 & .871
			& .702 &.071
			& .803 &.869 
			& .661 & .037 
			& .838 & .900
			& .755 & .049
			\\
			
			FocusDiffuser-P \cite{zhao2024focusdiffuser} &ECCV'24
			& - &-
			& - & -  
			& .881 & .939 
			& .851 & .042 
			& .875 & .939 
			& .809 &.020
			& .891 & .940 
			& .854 & .029
			\\
			
			UGDNet \cite{10855518}  &TMM'25 
			& .922 & .970 
			& .892 & .018 
			& \textcolor{blue}{\textbf{.888}} &\textcolor{red}{\textbf{.942}} 
			& \textcolor{blue}{\textbf{.865}} & \second{.038}
			& .885 & .947 
			& .822 & .019
			& .895 & .943 
			& .862 & .028
			\\
			
			CamoDiffusion \cite{10834569} &TPAMI'25
			& - & - 
			& - & -  
			& .878 & .940
			& .853 & .042 
			& .881 & .944 
			& .814 & .020 
			& .893 & .942 
			& .859 & .029
			
			\\
			
			UpGen \cite{11131534} &TIP'25
			& .820 & .874 
			& .757 & .048  
			& .793 & .842
			& .737 & .089 
			& .802 & .852 
			& .703 & .053 
			& .836 & .883 
			& .788 & .054
			
			\\
			\noalign{\hrule height 1pt} 
			\multicolumn{18}{l}{\textit{Segment Anything Model for  Camouflaged Object Detection}}
			\\ \hline
			
			SAM\cite{SAM} &ICCV'23 
			& .767 & .776
			& .696 & .078  
			& .684 & .687
			& .606 & .132 
			& .783 & .798 
			& .701 & .050 
			& .767 & .776 
			& .696 & .078
			\\
			
			SAM-Adapter\cite{c4} &ICCVW'23
			& .896 & .919 
			& .824 & .033 
			& .847 & .873
			& .765 &.070
			& .883 & .918 
			& .801 & .025 
			& - &-
			&- & - 
			\\ \rowcolor[rgb]{0.9,0.9,0.9}
			
			SAM2-UNet\cite{xiong2024sam2} &ICCVW'25
			& .914 & .961 
			& .863 & .022 
			& .884 & .932 
			& .861 & .042 
			& .880 & .936 
			& .789 & .021 
			& .901 & .941 
			& .863 & .029 
			
			\\
			
			FGSANet \cite{10844993} &TMM'25
			& .915 & .969 
			& \textcolor{blue}{\textbf{.901}} & \textcolor{blue}{\textbf{.017}} 
			& .873 & .938 
			& .853 & .043 
			& .889 & .936 
			& \textcolor{red}{\textbf{.839}} & \second{.018} 
			& .893 & .934 
			& \textcolor{blue}{\textbf{.865}} & \textcolor{blue}{\textbf{.027}}
			\\ \rowcolor[rgb]{0.9,0.9,0.9}
			
			\cite{xiong2024sam2} w. Mona \cite{yin20255} &CVPR'25
			& .915 & .969 
			& \textcolor{blue}{\textbf{.901}} & \textcolor{blue}{\textbf{.017}} 
			& .873 & .938 
			& .853 & .043 
			& .889 & .936 
			& \textcolor{red}{\textbf{.839}} & \second{.018} 
			& .893 & .934 
			& \textcolor{blue}{\textbf{.865}} & \textcolor{blue}{\textbf{.027}}
			\\
			
			\noalign{\hrule height 1pt} \rowcolor[rgb]{0.9,0.9,0.9}
			\textbf{Ours} & -
			& \textcolor{blue}{\textbf{.930}} & .965 
			& \textcolor{blue}{\textbf{.901}} & \textcolor{red}{\textbf{.016}}
			& \textcolor{red}{\textbf{.901}} & \textcolor{red}{\textbf{.942}} 
			& \textcolor{red}{\textbf{.872}} & \textcolor{red}{\textbf{.035}} 
			&\textcolor{red}{\textbf{.897}} & \textcolor{blue}{\textbf{.948}} 
			& \textcolor{blue}{\textbf{.837}} & \textcolor{red}{\textbf{.017}} 
			& \textcolor{red}{\textbf{.908}} & \textcolor{blue}{\textbf{.945}} 
			& \textcolor{red}{\textbf{.873}} & \textcolor{red}{\textbf{.026}}
			\\
			\noalign{\hrule height 1pt}
	\end{tabular}}
\end{table*}
\begin{table*}
	\renewcommand\arraystretch{1.2}
	\footnotesize
	\centering
	\caption{Comparison with SOTA methods on mirror detection. The best and second results are in \textcolor{red}{\textbf{red}} and \textcolor{blue}{\textbf{blue}}, respectively.}
	\label{tab:Mirror_Detection}
	\setlength{\tabcolsep}{3.5pt}
	\scalebox{1.0}{\begin{tabular}{l|c|ccccc|ccccc}
			\noalign{\hrule height 1pt}
			\multirow{2}{*}{Method}  & \multirow{2}{*}{Publish} &  \multicolumn{5}{c|}{PMD} & \multicolumn{5}{c}{MSD} \\ \multicolumn{1}{l|}{} & \multicolumn{1}{l|}{}
			&IoU $\uparrow$ & Acc $\uparrow$ & $F_\beta $ $\uparrow$&MAE $\downarrow$ & BER $\downarrow$&IoU $\uparrow$ & Acc $\uparrow$ & $F_\beta $ $\uparrow$&MAE $\downarrow$ & BER $\downarrow$ \\
			\noalign{\hrule height 1pt}
			MirrorNet~\cite{yang2019my} &ICCV'19 &58.51&70.18&0.741&0.043&15.61&78.95&93.29&0.857&0.065&6.39\\
			PMDNet~\cite{lin2020progressive} &CVPR'20 &66.00&77.15&0.785&0.033&12.13&81.55&88.20&0.892&0.047&7.17\\
			VST~\cite{liu2021visual} &ICCV'21 &59.06&-&0.769&0.035&-&79.09&-&0.867&0.052&-\\
			SANet~\cite{guan2022learning} &CVPR'22 &66.84&74.98&0.795&0.071&13.44&79.85&86.69&0.877&0.054&8.31\\
			VCNet~\cite{tan2022mirror} &TPAMI'22 &64.02&69.26&0.811&0.035&15.68&80.08&85.53&0.897&0.048&8.43\\
			SATNet~\cite{huang2023symmetry} &AAAI'23 &69.38&77.22&\first{0.847}&0.025&11.93&85.41&\second{89.14}&\first{0.922}&\second{0.033}&\second{6.21}\\
			HetNet \cite{he2023efficient} &AAAI'23 &69.0&-&0.814&0.043&-&82.8&-&0.906&0.043&-\\
			CSFwinformer~\cite{xie2024csfwinformer} &TIP'24 &70.05&\second{78.27}&\second{0.838}&\second{0.024}&\second{11.41}&82.08&88.92&0.896&0.045&7.14\\
			DPRNet~\cite{zha2024dual} &TCSVT'24 &72.10&-&-&0.026&-&86.60&-&-&0.033&-\\
			M2SD \cite{10988826} &TCSVT'25 &69.77&-&84.6&0.032&-&87.11&-&93.6&0.032&-\\\rowcolor[rgb]{0.9,0.9,0.9}
			SAM2-UNet~\cite{xiong2024sam2} &ICCVW'25 &\second{72.80}&-&-&0.027&-&\second{91.80}&-&-&\second{0.027}&-\\\rowcolor[rgb]{0.9,0.9,0.9}
			\cite{xiong2024sam2} w. MONA \cite{yin20255} &CVPR'25 &63.2&76.53&0.810&0.039&11.97&78.6&87.2&0.829&0.053&7.8\\
			
			\noalign{\hrule height 1pt} \rowcolor[rgb]{0.9,0.9,0.9}
			Ours &-&\first{79.25}&\first{89.94}&0.826&\first{0.018}&\first{5.80}&\first{92.43}&\first{96.86}&\second{0.907}&\first{0.021}&\first{2.55} \\
			\noalign{\hrule height 1pt}
	\end{tabular}}
\end{table*}
\subsubsection{Quantitative Evaluation}The quantitative results of the proposed method on texture-insensitive tasks are presented in Tab. \ref{tab:Marine_Animal_Segmentation}–\ref{tab:sota_Camouflaged_Object_Detection}. Overall, SAM-based fine-tuning methods demonstrate strong performance. Our method consistently outperforms the baseline across these tasks and establishes new state-of-the-art (SOTA) results on several benchmarks. In addition, the baseline combined with MoNA-based fine-tuning shows limited robustness. While it outperforms the baseline on a few individual tasks, its overall performance is significantly lower. In contrast, our fine-tuning method consistently surpasses MoNA across all tasks.

Specifically, the results for marine animal segmentation are presented in Tab. \ref{tab:Marine_Animal_Segmentation}. It can be seen that SAM-based fine-tuning methods, such as MAS-SAM (IJCAI'24), Dual-SAM (CVPR’24), our baseline, and our proposed method, all achieve strong performance, surpassing task-specific designs. In particular, our method achieves the best results across all metrics on both public datasets, with notable improvements over the baseline.
SAM-based fine-tuning methods benefit from the strong representational capacity of vision foundation models, which enables them to adapt well to complex scenes. Our baseline further enhances this capacity by effectively leveraging the model’s multi-scale context-awareness, contributing to its solid overall performance. However, factors such as insufficient lighting, color distortion, image blur, low contrast, and high variability in target appearance, especially in texture and color, can limit the effectiveness of texture-biased representations. In contrast, our method introduces shape-biased representations, which effectively address these challenges and lead to more robust segmentation performance.

The results for shadow detection are shown in Tab. \ref{tab:Marine_Animal_Segmentation}. In this task, SAM-based fine-tuning methods generally perform poorly. This is primarily because shadows in the ISTD dataset often co-occur with complex natural textures, making the task particularly challenging. Existing task-specific methods typically achieve high performance by jointly training for shadow detection and removal.
Our baseline performs unsatisfactorily on this task, and its performance further degrades after MoNA-based fine-tuning. This is likely due to the inability of both approaches to overcome texture bias, with fine-tuning potentially exacerbating this issue. It is worth noting that EVPv1 and EVPv2 \cite{11197268} demonstrate strong performance, as they introduce object contour information in the frequency domain, which provides a degree of robustness against texture interference. However, relying solely on contour information is insufficient. Our method integrates both object boundaries and internal distributions, achieving the best overall performance.

The results for camouflaged object detection are shown in Tab. \ref{tab:sota_Camouflaged_Object_Detection}. Since this task requires strong scene understanding and the data consist of conventional natural images \cite{luo2023camouflaged}, fine-tuning methods based on vision foundation models achieve relatively strong performance.
Our baseline demonstrates strong performance, as it effectively leverages the model’s multi-scale capabilities and is built upon a U-shaped architecture, which has been shown in practice to be highly effective for camouflaged object detection \cite{11175541}. Notably, the MoNA-based fine-tuning method is one of the few that surpasses the baseline, suggesting that it offers enhanced contextual modeling, which is particularly critical for this task.
Our method achieves the best performance compared to both the baseline and the MoNA-based fine-tuning, and also outperforms a range of task-specific approaches as well as other vision foundation model fine-tuning methods. This is largely because, in camouflaged object detection, the texture and color of the target are often highly similar to the surrounding environment. As a result, texture bias can lead to incorrect predictions.
In contrast, our method introduces shape-biased representations. Unlike FGSANet~\cite{10844993}, which captures object boundaries solely through frequency-domain information, our approach addresses the fact that boundaries in camouflaged scenarios are often blurred and indistinct from the background. Relying exclusively on boundary cues introduces uncertainty. By jointly modeling both boundary and internal feature distributions, our method produces more robust and reliable results.

The results for mirror detection are presented in Tab. \ref{tab:Mirror_Detection}. Mirror detection is a particularly challenging task because it requires distinguishing interference inside and outside the mirror. In this context, simple cues such as texture and color can often introduce noise that hampers detection performance. 
Overall, SAM-based fine-tuning methods do not demonstrate significant advantages over task-specific designs. This is primarily because simple fine-tuning fails to overcome the inherent texture bias within the model. Notably, the performance of MoNA fine-tuning is substantially lower than that of the baseline, confirming its limited generalization capability.
In contrast, our method outperforms the baseline on both datasets. In particular, on the highly challenging PMD dataset, our approach improves the IoU metric by 6.45 percentage points compared to the baseline. This improvement is comparable to the gains achieved between the current state-of-the-art (2025) and the state-of-the-art methods from 2020, highlighting the critical role of shape information.

\subsubsection{Qualitative Evaluation}
\begin{figure*}[h!]
	\centering
	\includegraphics[width=1\linewidth]{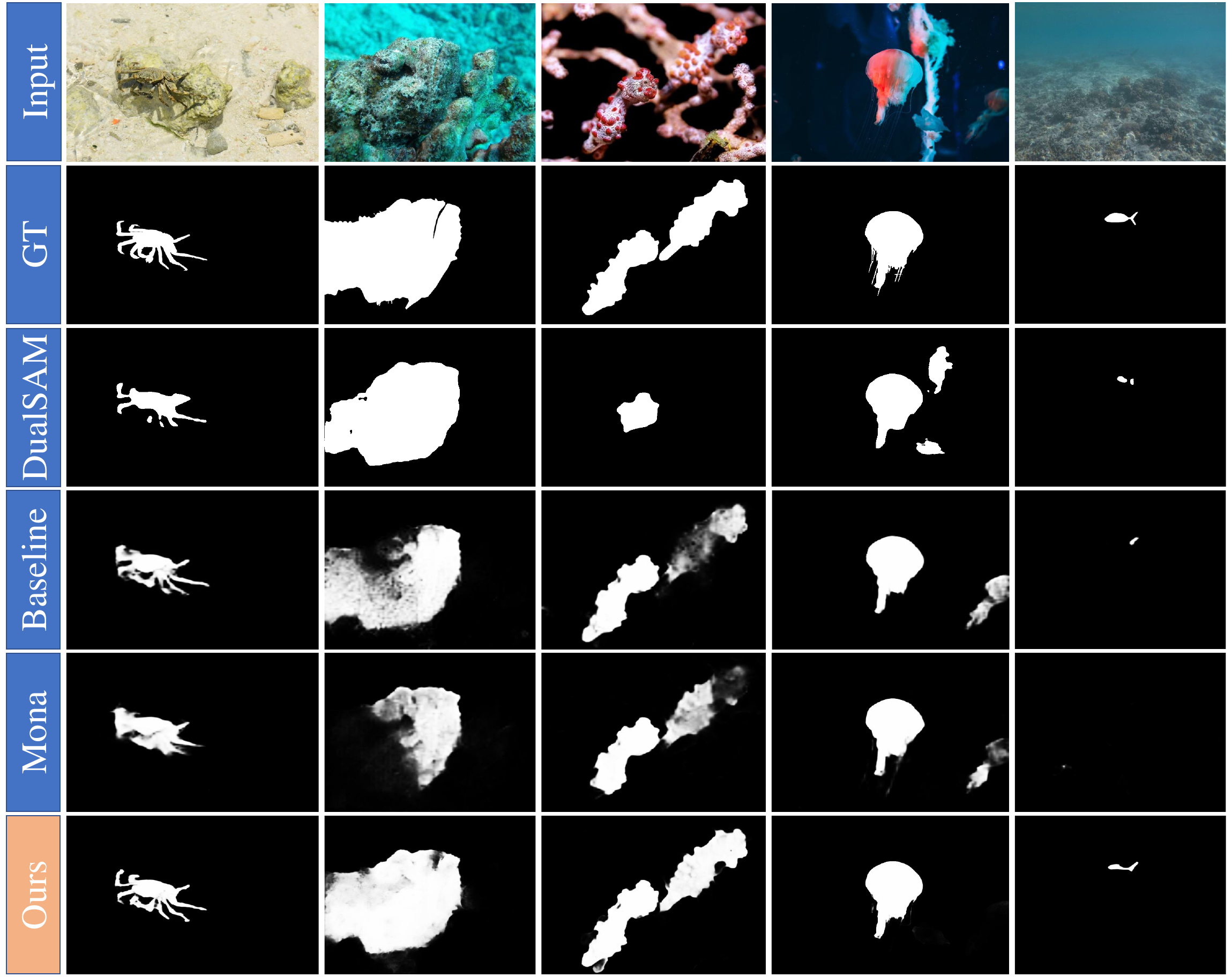}
	\caption{Qualitative comparison of segmentation results on the Marine Animal Segmentation task, demonstrating the performance of our method against Dual-SAM (CVPR'24), SAM2-UNet (ICCVW'25), Mona (CVPR'25).}
	\label{fig_Qualitative_Evaluation_Marine}
\end{figure*}
\begin{figure*}[h!]
	\centering
	\includegraphics[width=1\linewidth]{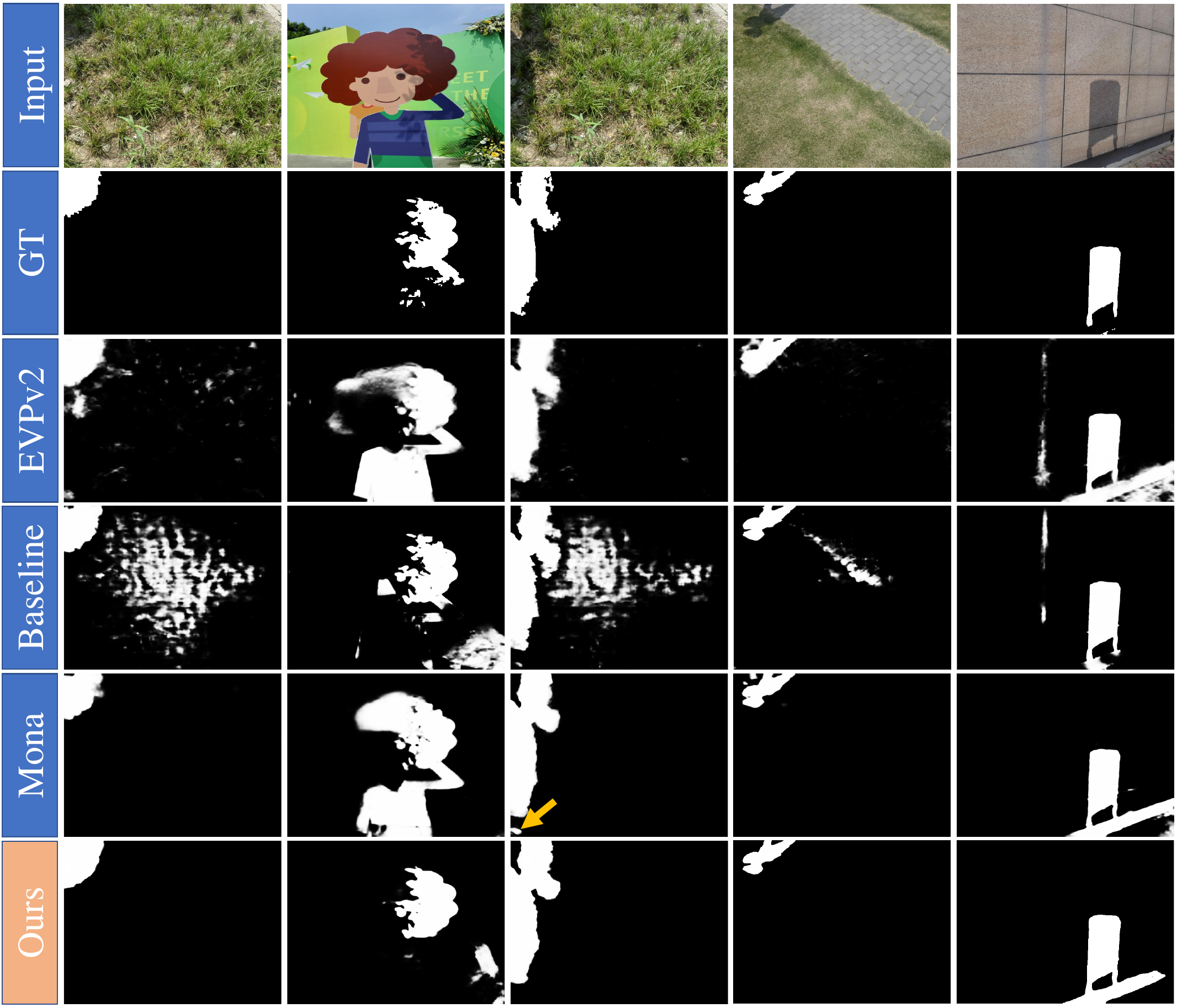}
	\caption{Qualitative comparison of segmentation results on the Shadow Detection task, demonstrating the performance of our method against EVPv2 (TPAMI'25), SAM2-UNet (ICCVW'25), Mona (CVPR'25).}
	\label{fig_Qualitative_Evaluation_ISTD}
\end{figure*}
\begin{figure*}[h!]
	\centering
	\includegraphics[width=1\linewidth]{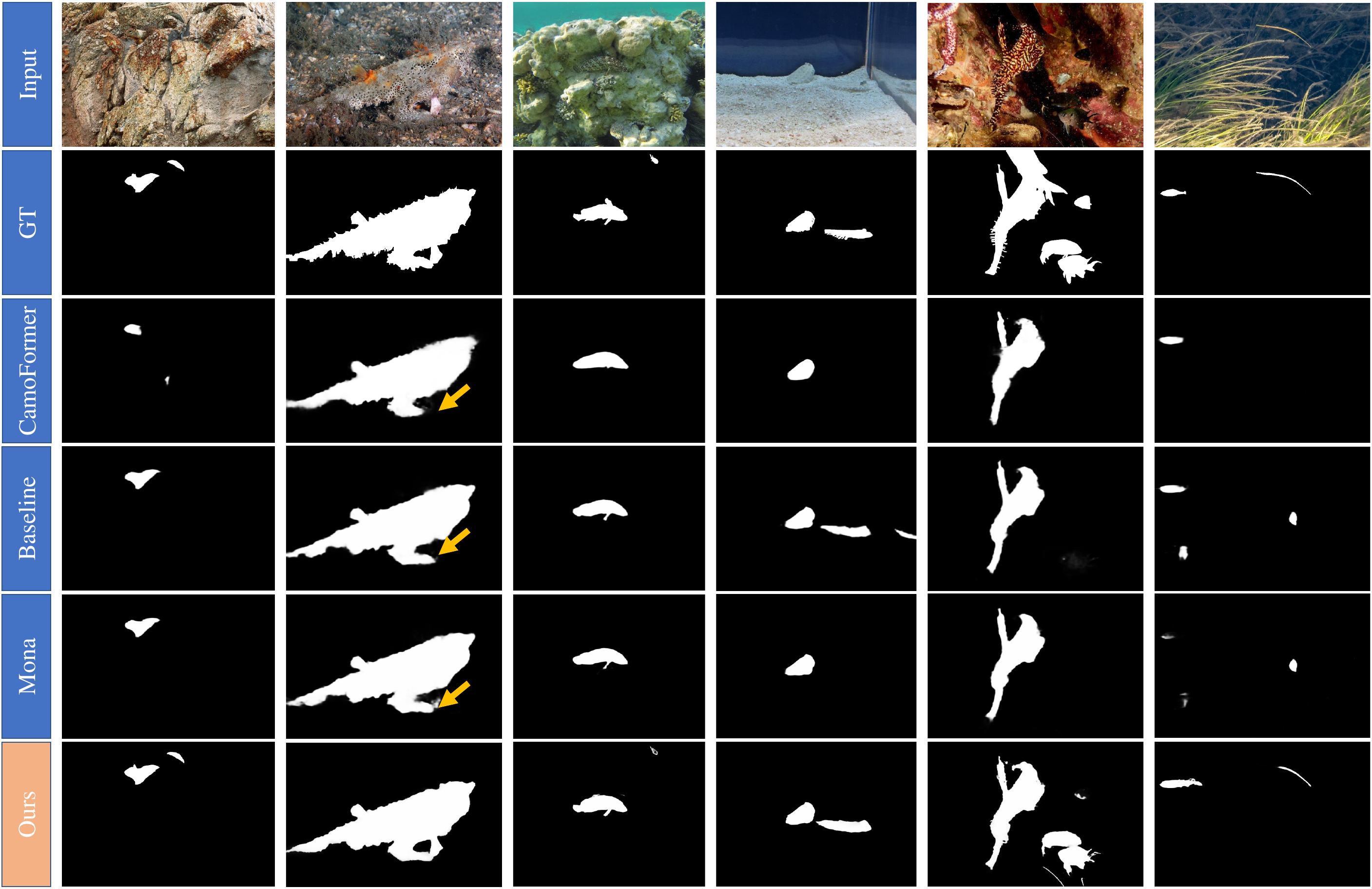}
	\caption{Qualitative comparison of segmentation results on the Camouflaged Object Detection task, demonstrating the performance of our method against CamoFormer (TPAMI'24), SAM2-UNet (ICCVW'25), Mona (CVPR'25).}
	\label{fig_Qualitative_Evaluation_COD}
\end{figure*}
\begin{figure*}[h!]
	\centering
	\includegraphics[width=1\linewidth]{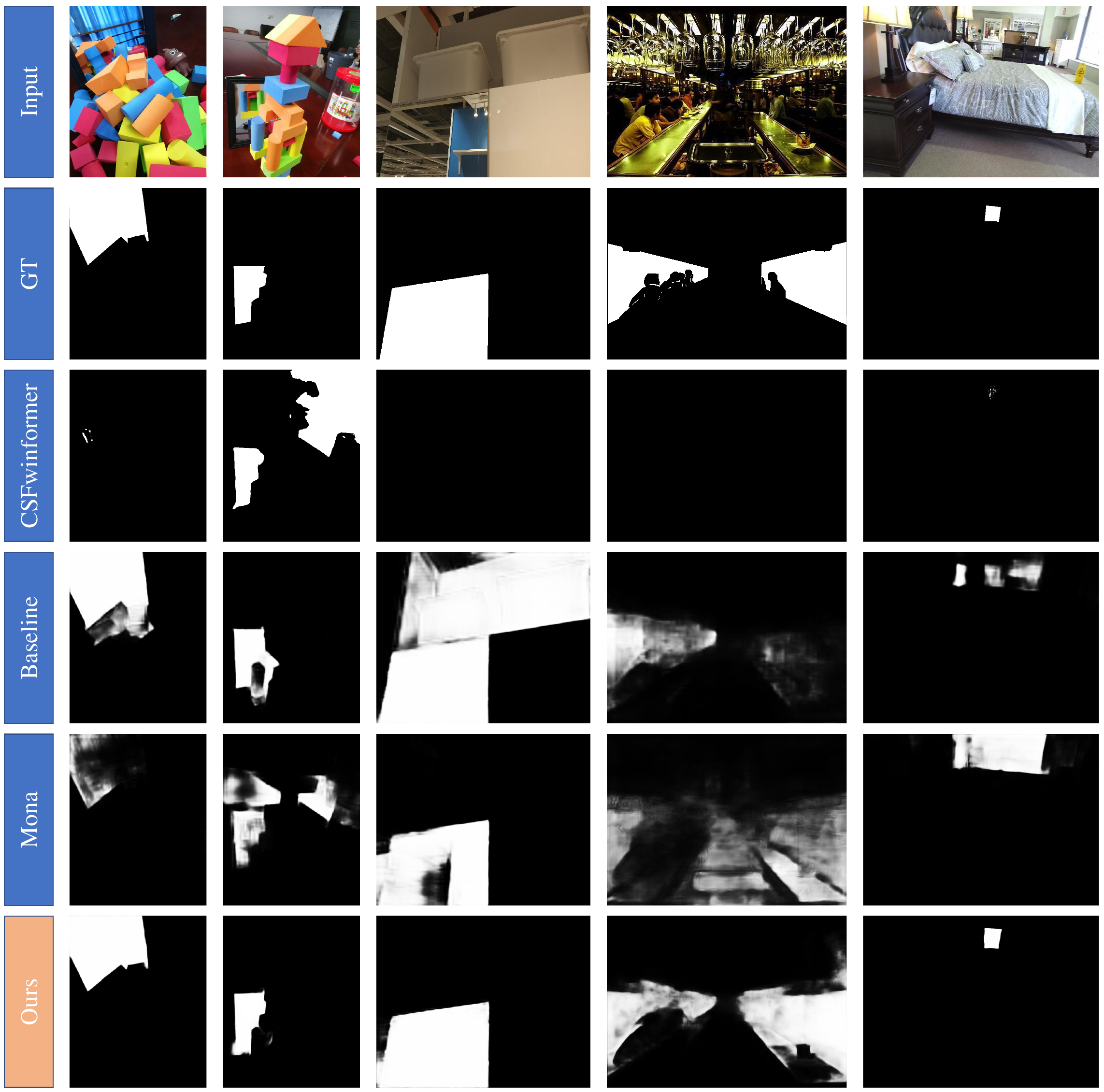}
	\caption{Qualitative comparison of segmentation results on the Mirror Detection task, demonstrating the performance of our method against CSFwinformer (TIP'24), SAM2-UNet (ICCVW'25), Mona (CVPR'25). Mirror detection is a highly challenging task, as it requires precise differentiation between objects inside and outside of mirrors, especially under complex conditions involving lighting variations, cluttered scenes, and the presence of people. Furthermore, the presence of other reflective surfaces such as glass adds additional ambiguity and complexity.}
	\label{fig_Qualitative_Evaluation_mirror}
\end{figure*}
As shown in Fig. \ref{fig_Qualitative_Evaluation_Marine}, our method achieves superior qualitative results on marine animal segmentation compared to the baseline. It demonstrates enhanced capability in perceiving fine-grained details, primarily due to the introduction of shape-biased representations. Furthermore, the shape bias contributes to improved robustness against texture interference, enabling our model to better suppress background regions even when the foreground and background share similar visual characteristics. As a result, the model exhibits increased sensitivity to target structures.

As illustrated in Fig. \ref{fig_Qualitative_Evaluation_ISTD}, our method demonstrates superior qualitative performance in the shadow detection task. In such scenarios, shadows are often closely coupled with complex scene content, where background textures tend to introduce a large number of false positives in conventional approaches. By incorporating shape-biased representations and focusing not only on object contours but also on internal structural distributions, our method effectively suppresses false alarms. As a result, it outperforms the EVPv2 method and significantly reduces the false detections present in the baseline.

As shown in Fig. \ref{fig_Qualitative_Evaluation_COD}, our method exhibits stronger perception of small camouflaged targets. This improvement is attributed to the introduction of shape-biased representations, which enable the model to perform fine-grained comparisons between target structures and background textures. As a result, our approach not only enhances the detection of small objects but also improves the segmentation precision of larger targets.

As shown in Fig. \ref{fig_Qualitative_Evaluation_mirror}, mirror detection is a highly challenging task, as it requires precise differentiation between objects inside and outside of mirrors, especially under complex conditions involving lighting variations, cluttered scenes, and the presence of people. The mirror regions in an image can range from very small areas to dominant portions of the scene. Furthermore, the presence of other reflective surfaces such as glass adds additional ambiguity and complexity. Our method achieves strong performance under these challenging conditions, maintaining robustness even in scenes that are difficult to distinguish by human perception. It significantly outperforms the baseline, and notably, the task-specific CSFwinformer (TIP'24) fails to function reliably in such scenarios.

\subsection{Texture-dependent Tasks} \label{Sec:De}
\begin{table*}[h!]
	\renewcommand\arraystretch{1.2}
	\footnotesize
	\centering
	\caption{Comparison with SOTA methods on salient object detection. The best and second results are in \textcolor{red}{\textbf{red}} and \textcolor{blue}{\textbf{blue}}, respectively.}
	\label{tab:Salient_Object_Detection}
	\setlength{\tabcolsep}{3.5pt}
	\scalebox{0.9}{\begin{tabular}{l|cccc|cccc|cccc|cccc|cccc}
			\noalign{\hrule height 1pt}
			\multirow{2}{*}{Method}  &  \multicolumn{4}{c|}{DUTS-TE (N=5019)} & \multicolumn{4}{c|}{DUT-OMRON (N=5618)}   & \multicolumn{4}{c|}{HKU-IS (N=4447)}   & \multicolumn{4}{c|}{ECSSD (N=1000)} & \multicolumn{4}{c}{PASCAL-S (N=850)} \\ \multicolumn{1}{l|}{} 
			& $S_\alpha \uparrow$ & $E_\phi$ $\uparrow$ & $F_\beta^m$ $\uparrow$ & $M \downarrow$  & $S_\alpha \uparrow$ & $E_\phi$ $\uparrow$ & $F_\beta^m$ $\uparrow$ & $M \downarrow$ & $S_\alpha \uparrow$ & $E_\phi$ $\uparrow$ & $F_\beta^m$ $\uparrow$ & $M \downarrow$  & $S_\alpha \uparrow$ & $E_\phi$ $\uparrow$ & $F_\beta^m$ $\uparrow$ &$M \downarrow$  & $S_\alpha \uparrow$ & $E_\phi$ $\uparrow$ & $F_\beta^m$ $\uparrow$ & $M \downarrow$ \\
			\noalign{\hrule height 1pt}

			PDRNet$_{22}$~\cite{9745960} & .877 & - & .845 & .035 & .846 & - & .756 & .052 & .924 & - & .911 & .027 & .927 & - & .920 & .032 & .865 & - & .816 & .061 \\
			TCRNet$_{23}$~\cite{9861608} & .880 & - & .842 & .034 & .843 & - & .748 & .054 & .923 & - & .908 & .026 & .928 & - & .917 & .031 & .865 & - & .817 & .059 \\
			BBRF$_{23}$~\cite{ma2023boosting} & .908 & .927 & .916 & .025 & .855 & .887 & .843 & .042 & .935 & .965 & \textcolor{blue}{\textbf{.958}} & .020 & .939 & .934 & \second{.963} & .022 & .871 & .867 & .891 & .049\\
			MENet$_{23}$~\cite{wang2023pixels} & .905 & .938 & .918 & .028 & .850 & .871 & .845 & .045 & .927 & .960 & .951 & .023 & .928 & .951 & .957 & .021 & .872 & .910 & \textcolor{blue}{\textbf{.897}} & .053\\
			ICON$_{23}$ \cite{9789317}  & .917 & - & .886 & .025 & .869 & - & .804 & .043 & .935 & - & .925 & .022 & .941 & - & .936 & .023 & .885 & - & .854 & .048\\
			PoolNet+$_{23}$ \cite{9669083}  & .894 & - & - & .039 & .831 & - & - & .056 & .941 & - & - & .034 & .949 & - & - & .040 & .879 & - & - & .068\\
			EVPv1$_{23}$~\cite{liu2023explicit_CVPR} & .913 & .947 & .923 & .026 & .862 & .894 & \textcolor{blue}{\textbf{.858}} & .046 & .931 & .961 & .952 & .024 & .935 & .957 & .960 & .027 & .878 & .917 & .872 & .054\\
			EVPv2$_{23}$~\cite{liu2023explicit} & .915 & .948 & .923 & .027 & .862 & .895 & .857 & .047 & .932 & .963 & .953 & .023 & .935 & .957 & .958 & .028 & .879 & .917 & .869 & .053\\
			MDSAM$_{24}$~\cite{gao2024multi} & .920 & .949 & \textcolor{red}{\textbf{.937}} & .024 & .878 & .910 & \textcolor{red}{\textbf{.887}} & .039 & .941 & .969 & \textcolor{red}{\textbf{.963}} & \textcolor{blue}{\textbf{.019}} & .948 & .967 & \textcolor{red}{\textbf{.974}} & .021 & .882 & .917 & \textcolor{red}{\textbf{.907}} & .052\\
			
			ADMNet$_{24}$~\cite{10555313} & .849 & .893 & .813 & .052 & .826 & .869 & .763 & .058 & .901 & .946 & .906 & .036 & .900 & .933 & .909 & .049 & .815 & .862 & .796 & .088\\
			VST++$_{24}$ \cite{10497889} & .923 & - & - & .024 & .871 & - & - & .044 & .941 & - & - & .021 & \second{.950} & - & - & .021 & .889 & - & - & .056\\ 
			
			OURS$_{24}$ \cite{10562209} & .881 & .927 & .824 & .039 & .836 & .873 & .741 & .054 & .920 & .961 & .899 & .029 & .920 & .953 & .902 & .038 & .853 & .897 & .805 & .068\\\rowcolor[rgb]{0.9,0.9,0.9}
			
			SAM2-UNet$_{24}$~\cite{xiong2024sam2} & .934 & .959 & - & \textcolor{blue}{\textbf{.020}} & .884 & .912 & - & .039 & .941 & \textcolor{red}{\textbf{.971}} & - & \textcolor{blue}{\textbf{.019}} & \textcolor{blue}{\textbf{.950}} & \textcolor{blue}{\textbf{.970}} & - & .020 & .894 & .931 & - & \textcolor{blue}{\textbf{.043}}\\
			
			SENet$_{25}$ \cite{10844040} & .932 & .959 & \second{.931} & \second{.020} & .878 & .907 & .839 & .039 & .939 & .968 & .949 & .020 & .946 & .966 & .958 & .022 & .887 & .921 & .882 & .047\\
			
			SDNet$_{25}$ \cite{11072100} & - & - & .771 & .078 & - & - & .726 & .084 & - & - & .883 & .057 & - & - & .897 & .063 & - & - & .796 & .057\\
			
			DiffSOD$_{25}$ \cite{11031107} & .913 & .938 & - & .026 & .865 & .891 & - & .043 & .934 & .951 & - & .025 & .941 & .958 & - & .024 & .878 & .912 & - & .051\\
			
			UniSOD$_{25}$ \cite{11082344}  & .925 & .938 & .906 & .021 & .876 & .910 & .822 & \second{.037} & .940 & .969 & .939 & \first{.018} & \second{.950} & .941 & .953 & \first{.017} & .889 & .887 & .875 & .044\\\rowcolor[rgb]{0.9,0.9,0.9}
			
			\cite{xiong2024sam2} w. Mona$_{25}$ \cite{yin20255}  & \second{.935} & \second{.960} & .912 & \second{.020} & \second{.886} & \second{.913} & .831 & .038 & \second{.943} & \second{.970} & .935 & \second{.019} & \second{.950} & .969 & .947 & .019 & \second{.895} & \first{.933} & .869 & \second{.043}\\
			
			\noalign{\hrule height 1pt} \rowcolor[rgb]{0.9,0.9,0.9}
			Ours  & \textcolor{red}{\textbf{.936}} & \textcolor{red}{\textbf{.961}} & .920 & \textcolor{red}{\textbf{.019}} & \textcolor{red}{\textbf{.893}} & \textcolor{red}{\textbf{.923}} & .846 & \textcolor{red}{\textbf{.033}} & \textcolor{red}{\textbf{.944}} & \textcolor{red}{\textbf{.971}} & .941 & \textcolor{red}{\textbf{.018}} & \textcolor{red}{\textbf{.954}} & \textcolor{red}{\textbf{.971}} & .951 & \second{.018} & \textcolor{red}{\textbf{.897}} & \textcolor{blue}{\textbf{.932}} & .872 & \textcolor{red}{\textbf{.042}}\\
			\noalign{\hrule height 1pt}
	\end{tabular}}
\end{table*}
\subsubsection{Quantitative Evaluation}
The results for salient object detection are shown in Tab. \ref{tab:Salient_Object_Detection}. Salient object detection is a task that heavily relies on texture, color, and other related visual cues. 
Overall, SAM-based fine-tuning methods significantly outperform task-specific designs. This can be attributed to two main reasons. First, salient object detection requires strong scene understanding \cite{lu2019see}, which is a strength of large-scale pretrained vision foundation models. Second, the inherent texture bias within these models makes them particularly adept at leveraging texture information for prediction, aligning well with the characteristics of salient object detection tasks.
On the salient object detection task, MoNA-based fine-tuning outperforms the baseline, primarily due to its enhanced context-awareness. However, it still underperforms compared to our method. This is because our approach introduces shape bias into the representation. Notably, prior studies  \cite{lee2022improving, shi2020informative, theodoridis2022trapped, he2023shift, zhang2023understanding} have suggested that shape perception and texture perception are both essential components of the human visual system. By integrating these two perceptual cues, the human visual system is able to achieve robust recognition in complex visual environments. Our method draws inspiration from this synergy, resulting in improved performance.
\subsubsection{Qualitative Evaluation}
\begin{figure*}[h!]
	\centering
	\includegraphics[width=1\linewidth]{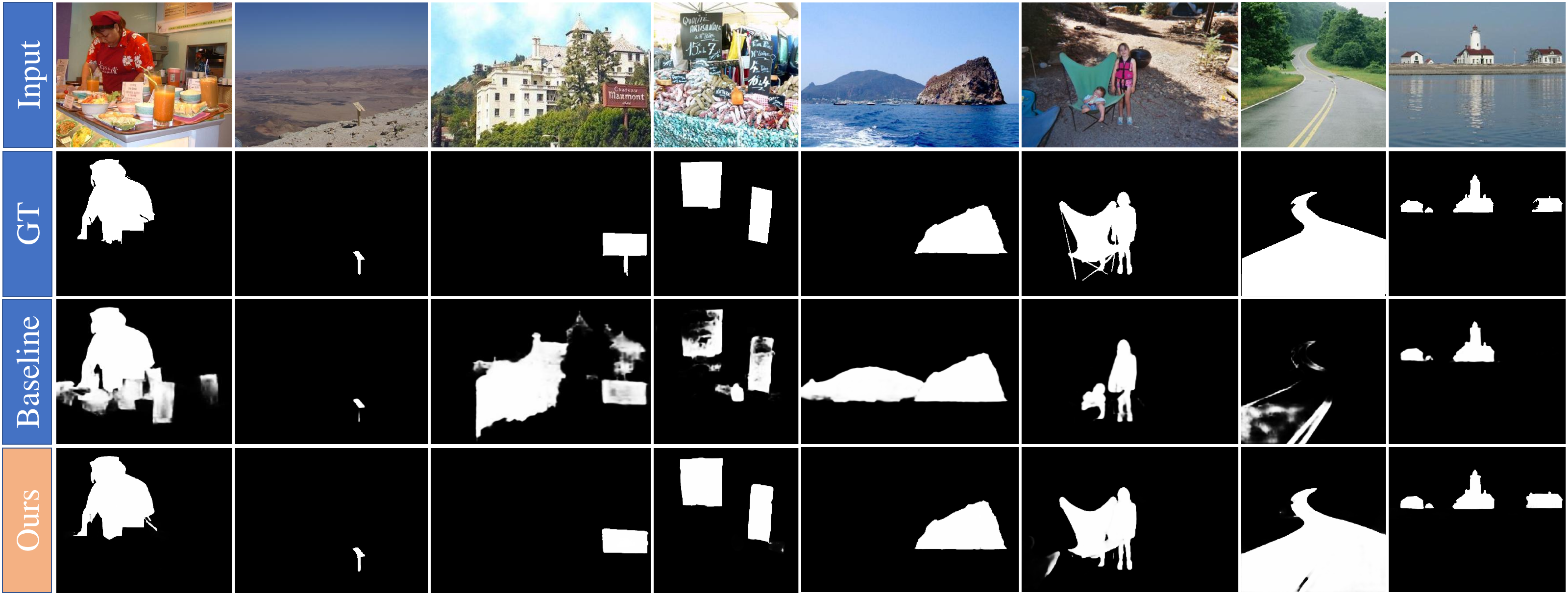}
	\caption{Qualitative comparison of segmentation results on the Salient Object Detection task.}
	\label{fig_Qualitative_Evaluation_SOD}
\end{figure*}
Qualitative results of salient object detection are presented in Fig. \ref{fig_Qualitative_Evaluation_SOD}, where we compare the baseline with our proposed method. It can be observed that our approach achieves superior performance in preserving fine details compared to the baseline. This improvement can be attributed to our method's focus on both object contours and internal feature distributions, enabling more precise handling of salient regions. Moreover, the side-branch subnetwork takes features from the ViT backbone as input, which already encode rich texture information. As a result, the side branch functions to further refine the response based on these texture-aware features, preventing any degradation in performance for tasks that are sensitive to texture details.

\subsection{Extended Discussion on Infrared Small Target Detection} \label{Sec:IR}
\subsubsection{Cross-dataset Generalization Performance}
\begin{figure}[h!]
	\centering
	\includegraphics[width=1\linewidth]{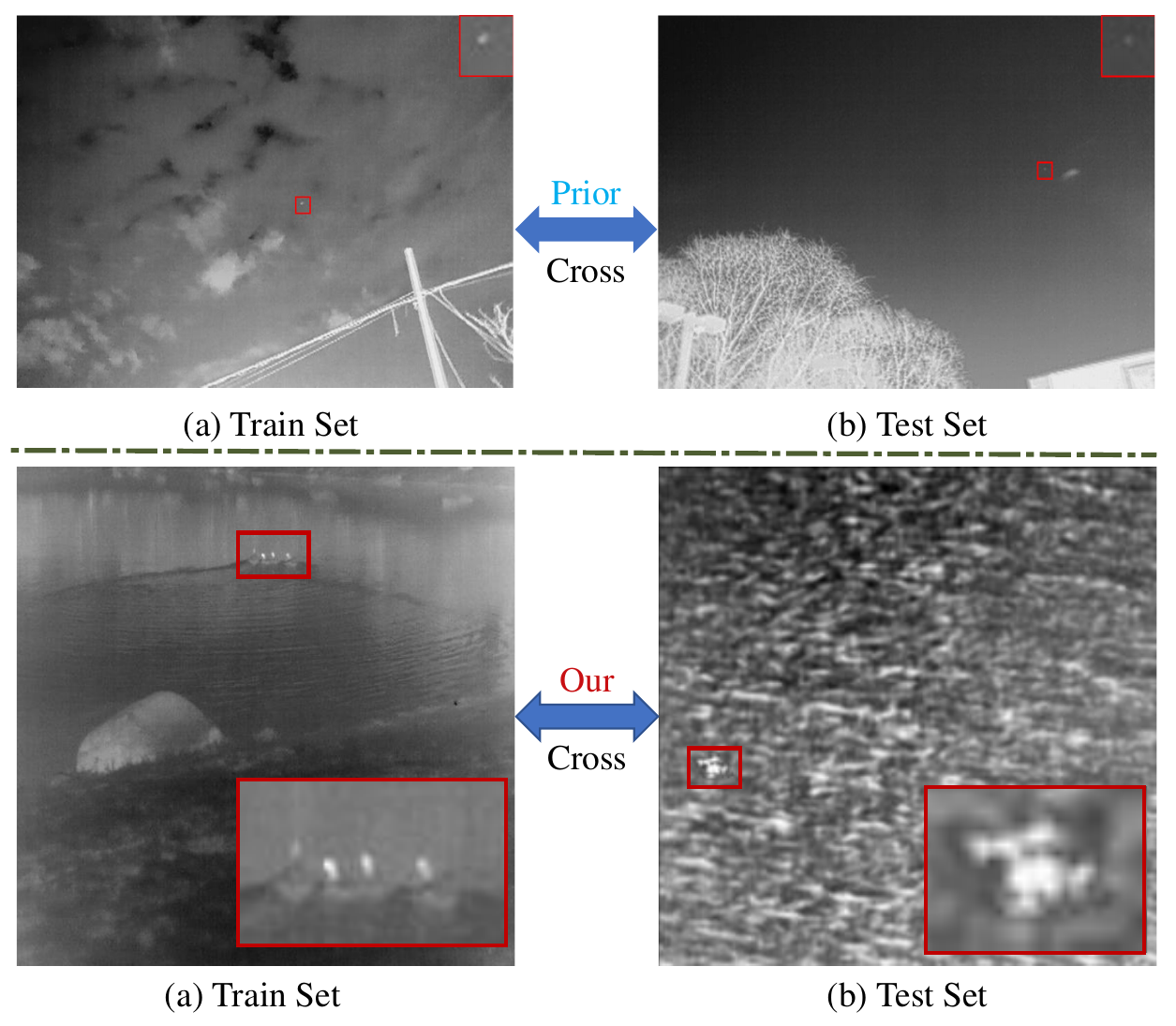}
	\caption{Cross-dataset generalization evaluation. Comparing the upper and lower subfigures, we observe that previous cross-dataset evaluations typically use datasets with similar target characteristics and scene contexts, which substantially lowers the requirements for model generalization. In contrast, our experimental setting involves a significant domain gap between the training and testing datasets, both in terms of target properties and background complexity. Specifically, the training set includes targets with pronounced scale variation and diverse object types such as humans and UAVs, whereas the test set contains only UAVs, which exhibit certain shape characteristics rather than simple point-like structures.}
	\label{fig_Cross_Qualitative_Evaluation}
\end{figure}
To further demonstrate the effectiveness of the proposed method, we conduct cross-dataset evaluations. It is worth noting that, as shown in Fig. \ref{fig_Cross_Qualitative_Evaluation}, unlike many prior studies \cite{RDIAN,DNANet,CSRNet} where the source and target datasets used for cross-dataset validation share similar target characteristics, our setting involves training and testing on datasets that differ significantly in terms of target properties and scene conditions.
This challenging setup is designed to better assess the model's generalization capability across diverse infrared scenarios.
\begin{table}[]
	\renewcommand\arraystretch{1.2}
	\footnotesize
	\centering
	\caption{Cross-dataset evaluation results. The model is trained on the IRSTD-1K dataset and tested on the NUDT-SIRSTD dataset. The metrics considered include IoU ($10^{-2}$), $\text{nIoU}$ ($10^{-2}$), $P_d$ ($10^{-2}$), $F_a$ ($10^{-6}$).}
	\label{tab:cross}
	\begin{tabular}{@{}c|c|c|cccc@{}}
		\toprule
		Methods &Train    & Test       & IoU   & nIoU  & $P_d$   & $F_a$   \\ \midrule
		\multicolumn{7}{l}{\textit{Deep Learning Methods}}  \\
		\hline
		Ours &IRSTD-1k & NUDT & \first{57.31} & \first{59.37} & \first{88.84} & 49.98 \\
		SAM-SPL \cite{11172325} &IRSTD-1k & NUDT & 19.94  & 28.00 & 52.86  & 271.3 \\ \hline
		\multicolumn{7}{l}{\textit{Model-Driven Methods}}  \\
		\hline
		NWMTH \cite{bai2010analysis}  & $\textemdash$ & $\textemdash$ & 11.72 & 10.26 & 52.71 & 46.81 \\
		ILCM \cite{GRSL2014ILCM}  & $\textemdash$ & $\textemdash$  & 12.84 & 13.69 & 27.62 & \second{43.21} \\
		MPCM \cite{PR16MPCM}  & $\textemdash$ & $\textemdash$ & 9.52  & 9.96  & 43.8  & 79.22 \\
		WLDM \cite{deng2016small}   & $\textemdash$ & $\textemdash$ & 11.04 & 21.03 & 12.69 & 78.26 \\
		RLCM \cite{GRSL2018RLCM}    & $\textemdash$  & $\textemdash$ & 18.37 & 18.11 & \second{79.51} & 67.13 \\
		FKRW \cite{qin2019infrared}  & $\textemdash$  & $\textemdash$  & 12.67 & 21.73 & \second{79.51} & 67.13 \\
		TLLCM \cite{han2019local}  & $\textemdash$  & $\textemdash$  & 11.08 & 23.82 & 77.97 & 67.26 \\
		GSWLCM  \cite{qiu2022global}  & $\textemdash$  & $\textemdash$ & 10.82 & 18.71 & 67.32 & 53.31 \\
		IPI \cite{gao2013infrared}   & $\textemdash$ & $\textemdash$   & 28.63 & 38.18 & 74.49 & \first{41.23} \\
		RIPT \cite{dai2017reweighted}  & $\textemdash$ & $\textemdash$  & \second{29.17} & 36.12 & 91.85 & 344.3 \\
		NRAM \cite{zhang2018infrared}   & $\textemdash$  & $\textemdash$  & 12.08 & 18.61 & 72.58 & 84.77 \\
		NOLC \cite{zhang2019infrared1}  & $\textemdash$  & $\textemdash$  & 23.87 & 34.9  & 85.47 & 58.2  \\
		PSTNN \cite{zhang2019infrared}  & $\textemdash$ & $\textemdash$   & 27.72 & \second{39.8}  & 66.13 & 44.17 \\
		\bottomrule
	\end{tabular}
\end{table}

The quantitative results are shown in Tab. \ref{tab:cross}. It can be observed that SAM-SPL \cite{11172325}, which is specifically designed for infrared small target detection, performs poorly. This is primarily because it only introduces a self-prompting module without modifying the model’s inherent bias. When there is a large domain gap between the training and testing datasets, such a design fails to adapt effectively.
In contrast, our method achieves significantly better performance. Although there is some performance drop due to the domain shift, it remains far superior to SAM-SPL and even outperforms several model-driven methods specifically tailored for infrared small target detection. These results highlight the potential of the proposed method, especially in scenarios where target domain data are scarce or difficult to obtain, making it a promising alternative to traditional model-driven approaches.

\subsubsection{Inference Speed Comparison}
To comprehensively evaluate the effectiveness of our method, we compare the inference speed with several representative approaches, including both traditional model-driven methods and deep learning-based methods. For deep learning methods, the input resolution is set to 512×512 with a batch size of 1, and inference is conducted on an NVIDIA A100 GPU. Model-driven methods are executed on CPU using the original hyperparameter settings reported in their respective papers. The quantitative results are summarized in Tab. \ref{tab:fps}.
\begin{table}[]
	\renewcommand\arraystretch{1.2}
	\footnotesize
	\centering
	\caption{Time Consumption of the Compared typical Methods (FPS)}
	\label{tab:fps}
	\setlength{\tabcolsep}{3.5pt}
	\begin{tabular}{cccccccc}
		\toprule
		IPI  & RIPT & PSTNN & DNANet & UIUNet & MTUNet & MSHNet & Ours  \\ \midrule
		0.3   & 1.5  & 4.4   & 4.56   & \second{5.8}   & 4.125  & 4.375   & \first{13.5} \\ \bottomrule
	\end{tabular}
\end{table}

In practice, although the side network introduced by our method slightly reduces the inference speed compared to the baseline, it still outperforms a wide range of existing infrared small target detection algorithms in terms of efficiency. This is because many of these methods \cite{DNANet,UIUNet,MSHNet,MTUNet} rely on highly complex modules or dense skip connections to preserve small target information in deep layers, which significantly slows down inference. In contrast, our approach leverages the strong representational capacity of SAM, reducing the need for such elaborate architectural designs.
Model-driven methods specifically designed for infrared small target detection are often even slower. Most state-of-the-art approaches in this category are based on low-rank and sparse decomposition techniques, which rely on complex iterative optimization procedures and are therefore time-consuming. In contrast, our method demonstrates promising efficiency and scalability, making it a compelling alternative for real-world deployment.

\subsection{Future Work} \label{Sec:Fu}
The downstream tasks reported in this supplementary material do not involve high-resolution images. In fact, with the advancement of imaging devices, exploring image processing in high-resolution scenarios is a promising direction \cite{shi2025auxdet,xiong2025sam2}. However, as shown in Tab. \ref{tab:time}, our side-tuning strategy introduces additional overhead, which makes the current method less suitable for direct extension to high-resolution segmentation.

Moreover, the tasks presented in this work, including our primary focus on infrared small target detection, fall under binary segmentation \cite{7837719,9320524}. Existing methods often lack sensitivity to fine structural details in complex environments \cite{qin2022highly,pei2023unite}, whereas results from both the main paper and supplementary material suggest that our method is more effective in this aspect. Future evaluation on diverse binary segmentation tasks (\eg, Forgery Detection \cite{10050168}, Defocus Blur Detection \cite{10036382}, Remote Sensing Salient Object Detection \cite{zhang2025advancing}) may provide further understanding of the role of shape bias, which is a promising direction for exploration.

Finally, considering the integration of our method with low-level vision tasks offers a promising direction. Detection outputs can provide informative priors for image restoration \cite{chen2024robustsam}, where more accurate segmentation may guide more effective reconstruction \cite{8960640}. Exploring this interplay could open up valuable opportunities for future research.
\bibliographystyle{IEEEtran}
\bibliography{reference}